\documentclass[journal]{IEEEtran}

\usepackage[T1,T5]{fontenc}
\usepackage[utf8]{inputenc}
\usepackage{amsmath,amssymb,amsfonts}
\usepackage{algorithmic}
\usepackage{textcomp}
\usepackage[hyphens]{url}
\usepackage[hidelinks]{hyperref}
\usepackage[UKenglish,abbreviations]{foreign}
\usepackage[style=ieee,sorting=none]{biblatex}
\usepackage{booktabs}
\usepackage{graphbox}

\graphicspath{{./img/}}

\newcommand{\ia}{\textit{i.a\@.}}
\newcommand{\SectionRef}[1]{\hyperref[#1]{Section~\ref*{#1}}}
\newcommand{\citeAfterAuthors}[1]{\cite{#1}}
\newcommand{\citep}[1]{\cite{#1}}

\begin{document}
\title{Deep Learning-based Single Image Face Depth Data Enhancement}
\author{
Torsten Schlett,
Christian Rathgeb,
and
Christoph Busch%
\thanks{T. Schlett, C. Rathgeb and C. Busch are with the  da/sec - Biometrics and Internet Security Research Group, Hochschule Darmstadt, Germany, \{torsten.schlett, christian.rathgeb, christoph.busch\}@h-da.de}%
}

\maketitle

\begin{abstract}
Face recognition can benefit from the utilization of depth data captured using low-cost cameras, in particular for presentation attack detection purposes. Depth video output from these capture devices can however contain defects such as holes or general depth inaccuracies.
This work proposes a deep learning face depth enhancement method
in this context of facial biometrics, which adds a security aspect to the topic.
U-Net-like architectures are utilized, and the networks are compared against
hand-crafted enhancer types, as well as a similar depth enhancer network from related work trained for an adjacent application scenario.
All tested enhancer types exclusively use depth data as input, which differs from methods that enhance depth based on additional input data such as visible light color images.
Synthetic face depth ground truth images and degraded forms thereof are created with help of PRNet,
to train multiple deep learning enhancer models with different network sizes and training configurations.
Evaluations are carried out on the synthetic data,
on Kinect v1 images from the KinectFaceDB,
and on in-house RealSense D435 images.
These evaluations include an assessment of the falsification for occluded face depth input, which is relevant to biometric security.
The proposed deep learning enhancers yield noticeably better results than the tested preexisting enhancers, without overly falsifying depth data when non-face input is provided,
and are shown to reduce the error of a simple landmark-based PAD method.
\end{abstract}

\section{Introduction}

\begin{figure}[tb]
	\centering
	\includegraphics[width=\linewidth]{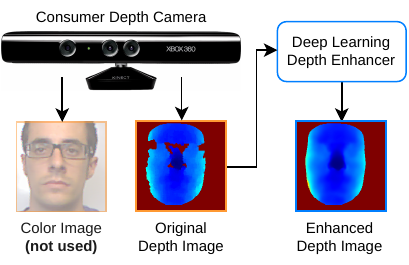}
	\vspace{-0.3cm}
	\caption{\label{fig:EnhancerPurpose} Deep learning face depth image enhancement concept. The color image (which is not used as input for the enhancement) and the original depth image stem from the KinectFaceDB \citep{14-KinectFaceDB, 14-KinectFaceDBSite} (``0011\_s1\_LightOn''), with the Kinect v1 therefrom representing a consumer depth camera. The proposed system was used to produce the enhanced depth image.
	Depth images are masked by the relevant face area for illustration.
	}\vspace{-0.2cm}
\end{figure}

Face recognition remains a challenging research area since several decades \citep{Zhao2003,Abate07,LiJain-HandbookOfFaceRecognition-2011}. More recently, developments in deep convolutional neural networks have shown impressive improvements in terms of recognition accuracy \citep{Parkhi2015,Schroff2015,Kawulok16a,Ranjan18a}. However, a variety of factors has been identified that can negatively impact recognition performance, including face image quality \citep{Galbally-Face-JRC34751SchengenInformationSystem-EuropeanUnion-2019,2020-FQA-Survey}. Focusing on 2D face imagery, a significant amount of research efforts has been devoted to face hallucination \citep{Baker00a,Liu2007} which represents a domain-specific super-resolution problem. In addition, many approaches to face completion \citep{GFC-CVPR-2017} have been proposed in the recent past. Published works which are mostly based on deep learning have been found advantageous in various face-related vision tasks, such as face detection and recognition \citep{GFC-CVPR-2017,Chen18aCV,Yu18a,Grm19,mathai2019doesgenerative}. With respect to 3D face data, similar concepts are expected to improve 3D face analysis, in particular for presentation attack detection (PAD), see Figure~\ref{fig:EnhancerPurpose}. 

\begin{figure}[tb]
	\centering
	\includegraphics[width=\linewidth]{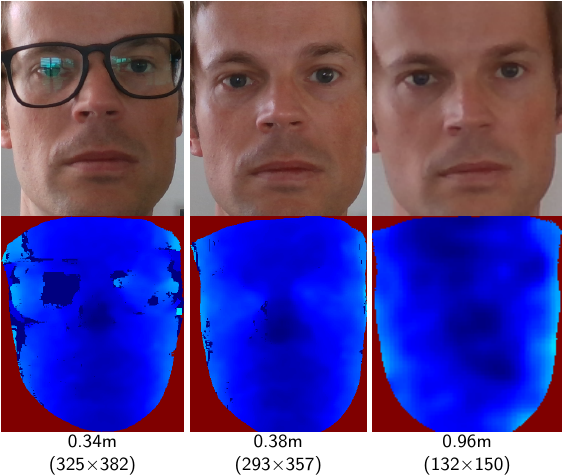}
	\caption{\label{fig:CamsRsDistEx0} RealSense D435 face depth images at two different distances 
	with resolution 1280×720 (second row). Depth quality and resolution markedly deteriorate, possibly making images recorded at about 1m or above unusable.
	Depth images are masked by the relevant face area for illustration.
	}\vspace{-1em}
\end{figure}

Inexpensive depth cameras
such as the Intel® RealSense™ Depth Camera D435 \citep{RealSenseD435}
can be used to obtain 3D face data.
See \autoref{fig:CamsRsDistEx0} for examples at different distances.
The low cost could facilitate the devices'
widespread usage as part of face biometric systems,
with \eg{} the RealSense D435 being priced at \$179 \citep{BuyRealSenseD435} at the time of this writing.
Another, arguably popular example for affordable ``consumer-grade'' depth cameras are the Kinect v1 and Kinect v2,
although both are discontinued products by now \citep{18-KinectDiscontinued, 19-KinectLiteratureReview},
with Microsoft offering the more expensive but still ``low-cost'' Azure Kinect at \$399 instead \citep{AzureKinectDK}. The mentioned devices are ``RGB-D'' cameras, meaning that they simultaneously record visible light (RGB) color data besides depth (D). The face depth images recorded by such cameras may contain undesirable holes (\ie areas of invalid data) due to \eg occlusion,
in addition to general depth value noise \citep{18-RealSenseDepthPostProcessing},
so an enhancement of the depth data may be desirable.

When it comes to image enhancement (for two-dimensional representations) via deep learning,
various promising results have been achieved for visible light color images in categories such as super-resolution \citep{19-DlSuperResolutionWoGan, 19-DlSuperResolutionSurvey}.
This raises the question whether a similar image-to-image deep learning approach could be used to effectively enhance face depth images,
which can technically be considered as
2D grayscale (\ie{} single channel) images,
with each pixel depicting a depth value.
This work aims to answer the question by creating face-specific deep learning depth image enhancers,
which are compared against existing general (\ie{} not face-specific) hand-crafted depth enhancers.
These also utilize pure depth image input without other helper data, as shown in \autoref{fig:EnhancerPurpose}.

\clearpage

Face recognition based on 3D or multi-modal 3D+2D
data has been shown to generally achieve higher accuracy than purely 2D
approaches \citep{T2D3DFRM2014}.
In terms of overall accuracy,
recent developments
in deep learning-based solutions have
surpassed other approaches \citep{DFRS2018, FRNMLTS2019},
whereby deep 3D face recognition developments are impeded by the lack of large-scale 3D datasets,
thus leading to the employment of 3D data synthesis \citep{DFRS2018}.
In addition,
deep learning-based face recognition approaches are known to be especially vulnerable to presentation attacks \citep{DV:SRFrPa-2018, DFRS2018}.
PAD approaches, also known as anti-spoofing, can thus
be employed as part of face recognition systems
to counter malicious disguises\slash mimicry \eg via face printouts or masks \citep{PADMFRS:CS-2017}.
Utilizing depth data as part of an anti-spoofing or face recognition system can increase the system's efficacy. 
In contrast to related works, which use RGB color images as helper data for depth enhancement, this work exclusively considers the depth data as input for depth enhancement. 
Since non-depth helper input is explicitly not taken into consideration by this work's enhancers,
a resulting advantage is that the depth enhancement itself obviously cannot be falsified through such data. This is essential in the context of depth-based face PAD where non-depth helper input, \ie RGB face images, might represent presentation attacks, \eg face printouts.

The contributions of this work can be summarized as follows:
\begin{itemize}
\item
We propose using deep learning to train face-specific depth enhancers,
with the intent to use the output for biometric systems such as depth-based face PAD.
To avoid strong detrimental falsification of the input data,
we exclusively use depth as input.
Multiple networks with U-Net-like architectures were created and trained from scratch.
\item
A synthesis of realistic face depth data (in ground truth \& degraded form) based on the PRNet algorithm \citep{18-PRNet} is used to
train the deep learning networks
and to conduct a quantitative enhancer comparison including preexisting hand-crafted depth enhancers
and a publicly available DDRNet depth denoiser network \citep{19-DDRNet-GitHub}.
\item
Qualitative and further quantitative evaluation using
real depth camera data,
which includes Kinect v1 images taken from the KinectFaceDB \citep{14-KinectFaceDB, 14-KinectFaceDBSite}
in addition to custom images recorded with the RealSense D435.
The quantitative real data evaluations examine
the level of input falsification for non-face input using the created deep learning enhancers,
and compare their impact on the hole percentage and roughness in contrast to hand-crafted depth enhancers.
\end{itemize}

The remainder of this work is organized as follows:
\SectionRef{sec:related-work} briefly summarizes related works,
\SectionRef{sec:proposed-system} describes the proposed system,
including the synthesis process,
\SectionRef{sec:experiments} presents the results of
the quantitative and qualitative evaluations.
Finally, in \SectionRef{sec:conclusion-and-future-work}, conclusions are drawn and suggestions for future work are given.

\section{Related Work}
\label{sec:related-work}

Different research fields are directly related to this work,
namely the overarching topic of depth enhancement,
as well as face depth synthesis.
The following subsection lists some of the related depth enhancement and depth estimation works,
whereby the former incorporate non-depth data for enhancement.
Subsequently, datasets and 3D face data synthesis approaches are briefly discussed.

\subsection{Depth Enhancement And Estimation}

Most related work in this subsection improves depth data, not necessarily of faces,
based on leveraging information provided by additional non-depth information such as RGB color images.
This stands in contrast to this work's deep learning enhancers that exclusively operate on face depth input,
and must not falsify input is case of presentation attacks on a biometric system,
which to our knowledge is a use case that has not been considered previously in the scientific literature.
Some of these related approaches do however have separable parts that purely use depth input.

A shading-based approach that relies on both depth and aligned RGB images
for consumer depth cameras  was proposed by Wu \etal{} \citeAfterAuthors{14-ShadingBasedRefinement}.
Kadambi \etal{} \citeAfterAuthors{15-Polarized3D} used three non-depth images taken using different polarization filters to enhance Kinect v2 depth images.
``DDRNet'' by Yan \etal{} \citeAfterAuthors{18-DDRNet} enhanced depth
with one pure depth denoising network,
and one subsequent refinement network that incorporates RGB color image input,
using a joint training strategy.
An exemplar-based approach for face depth super-resolution that incorporates higher resolution color image input,
which included the ability to denoise depth
but was not meant for hole-filling,
was presented by Yang \etal{} \citeAfterAuthors{Yang-FeatureGuidedFaceDepthSuperResolution-TCYB-2018}.
 Conditional generative adversarial networks (GANs) for simultaneous super-resolution of a low-quality color and depth input pair were used by Zhao \etal{} \citeAfterAuthors{zhaoColorDepthSRGAN2019}.
Li \etal{} \citeAfterAuthors{liHighQuality3DReconstruction2019} created a general depth super-resolution approach, which used surface normal, boundary and blur information estimated from a high resolution color image to facilitate depth hole-filling.
More recently,
Shabanov \etal{} \citeAfterAuthors{shabanovSelfsupervisedDepthDenoising2020} trained a simple U-Net-like network for single image depth denoising,
and optionally applied a subsequent ConvLSTM \citep{shiConvolutionalLSTMNetwork2015} to leverage temporal information.

Another related task is the completion of sparse depth images \eg{} for autonomous navigation:
Lu \etal{} \citeAfterAuthors{Lu_2020_CVPR}
used grayscale visible wavelength images as additional information during network training in their single image sparse depth completion approach.
A probabilistic normalized CNN for sparse depth completion was introduced by Eldesokey \etal{} \citeAfterAuthors{Eldesokey_2020_CVPR}.
Rossi \etal{} \citeAfterAuthors{Rossi_2020_CVPR} created an approach specialized for scenes than can be approximated by planar surfaces,
such as various human made structures.

Besides approaches that enhance depth data, with or without additional non-depth input,
there exist further approaches that estimate depth from color images without any depth input.
While using purely estimated depth alone for depth-based PAD would be questionable
due to the risk of generating highly inaccurate depth in case of an attack,
it could be utilized as a basis to enhance real depth input.
Depth estimation can also be used to generate training or testing data for an enhancement method,
and further works related specifically to 3D face synthesis are listed separately in \autoref{sec:3d-face-synthesis}.
Among recent works in this category,
Wang \etal{} \citeAfterAuthors{Wang_2019_CVPR} proposed a monocular approach including a U-shaped architecture similar to DispNet \citep{MIFDB16} with recurrent units to leverage temporal information for depth estimation,
Gur and Wolf \citeAfterAuthors{Gur_2019_CVPR} utilized focus cues in their architecture to estimate depth for single images,
and Poggi \etal{} \citeAfterAuthors{Poggi_2020_CVPR} examined uncertainty estimation as part of single image depth estimation.

The interested reader is referred to \citep{ListDepthEnhancement} for a list other related, albeit mostly non-face, depth enhancement works.

\subsection{Datasets}
\label{sec:datasets}

A recent study from Zhou and Xiao \citeAfterAuthors{3DFRS2018}
lists available 3D face datasets.
This work only considers depth data as input,
so the quality of other data such as color images in these datasets is not important.
The depth quality however needs to be sufficient to fulfill the role of ground truth images.
This means that the quality should at least be noticeably higher than that of the low-cost depth camera images (such as the RealSense D435 or one of the Kinect variants).
Optimally the ground truth should not have any depth defects such as noise or holes at all,
so even expensive high-quality scanner output might not be optimal without post-processing.
Imperfect counterparts comparable to the ground truth are required as well to train and test enhancers.
To enable direct comparisons, these should be recorded from approximately the same viewpoint with the same face orientation and positioning,
so preferably at the same time.
Since such ground truth/degraded image pairs are not available in the known datasets,
one could alternatively degrade the real ground truth images synthetically.

In addition,
the amount of images might become a bottleneck for the deep learning enhancer training.
Among the listed datasets in \citep{3DFRS2018},
the largest subject count is 888 for the ND2006 dataset, with 13\,450 images total.
Alternatively, BU-4DFE has the highest image count with 60\,600,
but these stem from only 101 subjects and may be especially similar since the data consists of 3D videos \citep{3DFRS2018, BU-4DFE}.
Also note that various datasets may contain images for different facial expressions per subject (\eg{} ND2006),
but not necessarily different head poses.

Some other datasets not covered by \citep{3DFRS2018} are \eg{}:
The ``KinectFaceDB'' (52 subjects, Kinect v1) \citep{14-KinectFaceDB},
the ``RGB-D Face database'' (31 subjects, Kinect v1) \citep{12-KinectRGBDFaceDatabase},
or the ``IAS-Lab RGB-D Face Dataset'' (26 subjects, Kinect v2) \citep{16-IASLabFaceDB}.
All three datasets have been recorded using Kinect variants,
so it can be assumed that the depth image quality is not sufficient to serve as ground truth images in this work,
besides the low subject counts.

To circumvent these real dataset issues,
this work fully synthesizes ground truth and degraded face depth image variants for the training of deep learning enhancers.
Experiments are carried out using both synthesized and real data.
Real data experiments use the KinectFaceDB and custom RealSense D435 recordings.

\subsection{3D Face Synthesis}
\label{sec:3d-face-synthesis}

A recent survey by Wang and Deng \citeAfterAuthors{DFRS2018} noted that
the number of scans and subjects in public 3D face databases is still limited,
hindering the development of 3D deep face recognition,
with synthesis being one option to obtain more 3D data.
Regarding PAD in particular, Atoum \etal{} \citeAfterAuthors{FAsPDCnn-2017} estimated depth maps from 2D face images.
Gilani and Mian \citeAfterAuthors{LFM3dSLS3dFR-2018} proposed a method to generate millions of detailed 3D faces by interpolating between facial identities and expression spaces from existing 3D data,
however their
used source dataset comprising 3D facial scans from 1\,785 individuals is proprietary
and their second source of 3D faces is a commercial software as well.
Similar in detail, but focused completely on the reconstruction,
Trần \etal{} \citeAfterAuthors{E3dFR:STO-2018} generated
3D data for single (occluded) 2D face images,
and the code was made publicly available \citep{E3dFR:STO-github}.
In contrast to these methods with detailed 3D face results,
Chinaev \etal{} \citeAfterAuthors{MF:3dFRECNNR-2018}
clearly produced less detailed output and the focus lies on attaining
sufficiently high computational efficiency for real-time mobile applications instead.
Most recently,
Baby \etal{} \citeAfterAuthors{Baby-FaceDepthEstimation-ACCTHPA-2020} proposed using a conditional GAN to generate face depth from color images,
with a U-Net architecture for the generator part.

Feng \etal{} \citeAfterAuthors{18-PRNet} presented
another real-time method
(when using a modern GPU such as their employed GTX 1080),
which also provides dense alignment of the generated 3D face data to the corresponding 2D input images.
The associated open source implementation \citep{PRNet-GitHub} is available and referred to as PRNet in this work, which uses it as part of the ground truth synthesis described in \autoref{sec:ground-truth-synthesis}.

\section{Proposed System}
\label{sec:proposed-system}

The proposed system can be subdivided intro three parts,
with each part building upon the previous part:
Firstly the face depth ground truth image synthesis,
secondly the face depth image degradation synthesis,
and lastly the actual deep learning depth enhancement network training.
Since the enhancement networks have a fixed input\slash output image resolution,
all synthesized images are generated in the same resolution as well.
We selected a spatial resolution of 256×256 pixels for this resolution,
which can be approximately comparable to the RealSense D435 face ROI resolution at a capture subject to capture device distance of about 0.5m.

Both ground truth images and degraded \ie{} ``improvable'' images are synthesized
to train and quantitatively test the proposed deep learning enhancer.
The overall synthesis process uses color face images as input data.
From these color images the ground truth depth images are synthesized.
Then each ground truth image can be synthetically degraded in a number of ways,
thus producing one or more degraded depth images directly corresponding to one ground truth image.

\begin{figure}[htb]
	\centering
	\includegraphics[width=\linewidth]{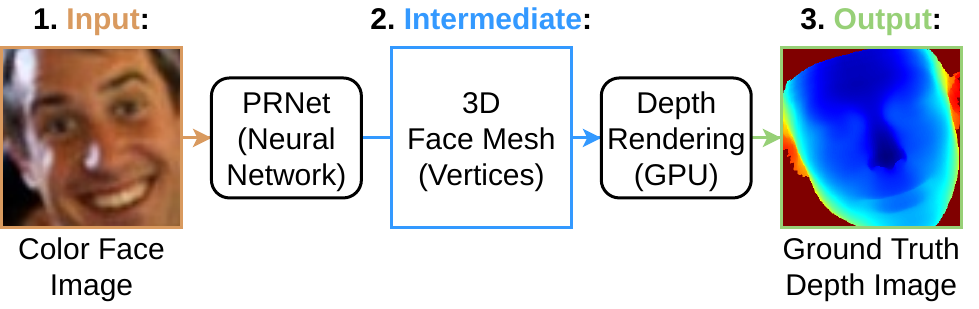}\vspace{-0.3cm}
	\caption{\label{fig:SynthGtProcess} The ground truth synthesis process: PRNet is applied to 2D face images in order to produce realistic depth data (example image taken from the MegaFace dataset).}\vspace{-1em}
\end{figure}

\subsection{Ground Truth Synthesis}
\label{sec:ground-truth-synthesis}

Synthesizing the ground truth depth images is the first, fundamental stage of the entire synthesis procedure;
the second and final stage being the subsequent degradation synthesis.
The implemented ground truth synthesis uses 2D visible light RGB color face images with arbitrary resolutions as input,
and produces 256×256 16bit ground truth face depth images as output,
meaning that each color input image results in exactly one ground truth output depth image.

Since the color images are only relevant as input for this synthesis stage,
there is no requirement for the synthesized ground truth images to properly align with
the face shapes depicted by the color input.
Instead, the synthesized ground truth images only have to depict
realistic and varied face depth data.
We use PRNet \citep{18-PRNet} for this synthesis,
which was shown to be more accurate than multiple prior methods
for 3D face alignment and reconstruction,
and better than the AFLW2000-3D \citep{Zhu-3dFaceAlignmentLargePoses-TPAMI-2019}
ground truth for face alignment in some cases.
Based on this evaluation and manual examination, we assume that the 3D face output of PRNet is sufficiently realistic to serve as ground truth here.
Note that PRNet does not generate facial details such as wrinkles,
but this is actually desirable since we enhance low-fidelity face depth data,
which means that generating higher levels of detail with an enhancer would be a detrimental falsification
for biometric purposes such as face depth PAD.

A filtered subset of the ``1M Disjoint Distractors'' set (``Tightly Cropped'' variant), belonging to the ``MF2 Training Dataset'', with ``MF'' standing for ``MegaFace'' \citep{17-MegaFace2, MegaFace2-Download}, was used to provide the color input data.
The full 1M Disjoint Distractors set contains one million color images.
$856\,129$ color images were left in the dataset after filtering out images with
identical SHA3-512 file hash results,
a resolution above 1024×1024 in either width or height (to remove a small percentage of outliers),
as well as overly deformed ground truth face output. Among the color image filter stages above,
``deformed output'' refers to PRNet 3D face mesh output that does not resemble a realistic face.
Such flawed output can occur \eg{} when the color input image quality is insufficient.
To discard these flawed input\slash output pairs,
428 well-formed 3D face output models were manually selected from a number of random samples
to determine acceptable value ranges for the model triangle side lengths.
These values were then used to filter all other models and their corresponding color input data.
This is possible because all PRNet face model output shares the same number of triangles in identical order. 
The discarded image counts are listed in \autoref{tab:SynthGtDiscarded}.

\begin{table}[htb]
	\caption{\label{tab:SynthGtDiscarded} The $1\,000\,000$ images of MF2's 1M Disjoint Distractor set were filtered in three subsequent stages to discard undesirable images.}\vspace{-0.0cm}
	\centering
	\begin{tabular}{l r}
		\toprule
		\textbf{Filter stages} & \textbf{Count} \\
		\midrule
		1. File duplicates (SHA3-512) & $213$ \\
		2. Resolution above 1024×1024 & $12\,087$ \\
		3. Deformed output & $131\,571$ \\
		\midrule
		Total discarded images: & $143\,871$ \\
		Remaining images: & $856\,129$ \\\bottomrule
	\end{tabular}
\end{table}

\begin{figure}[tb]
	\centering
	\includegraphics[width=\linewidth]{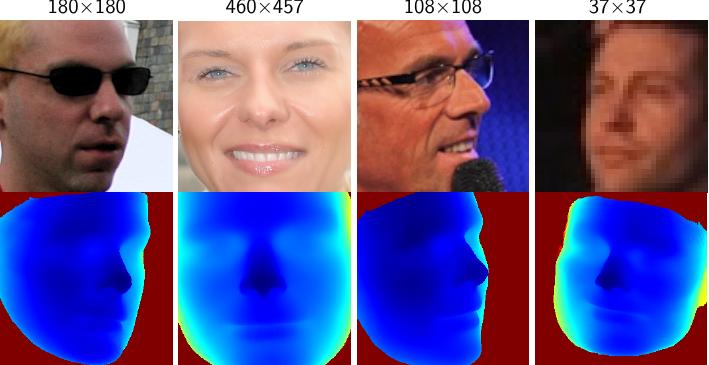}
	\caption{\label{fig:SynthGtOk} First row: Source color images with resolution labels. Second row: Colorized 256² synthetic ground truth output, revealing some alignment inaccuracies, which are however irrelevant since the face output remains realistic.}\vspace{-0.0cm}
\end{figure}

An overview of the actual ground truth synthesis process is illustrated in \autoref{fig:SynthGtProcess}.
It works by supplying one of the color face images as input for PRNet \citep{PRNet-GitHub},
the Tensorflow-based Python implementation provided by
Feng \etal{} \citeAfterAuthors{18-PRNet}.
At its core, PRNet is a convolutional neural network trained to output 3D-face-information-encoding ``UV position maps''
from 2D RGB color face images (scaled to 256×256) \citep{18-PRNet}.
Information in a UV position map can then be turned into (other) 3D data, such as a 3D mesh of the face, \ie{} a list of 3D vertices forming triangles.
PRNet inherently
aligns its output to the face of the color input image.
The alignment itself is not required by any subsequent steps of the system since the color image is not used,
but this means that the produced 3D output already does have a variety of orientations depending on the color image input.
\autoref{fig:SynthGtOk} shows some successful ground truth synthesis examples.

\subsection{Degradation Synthesis}

\begin{figure}[!tb]
	\centering
	\includegraphics[width=\linewidth]{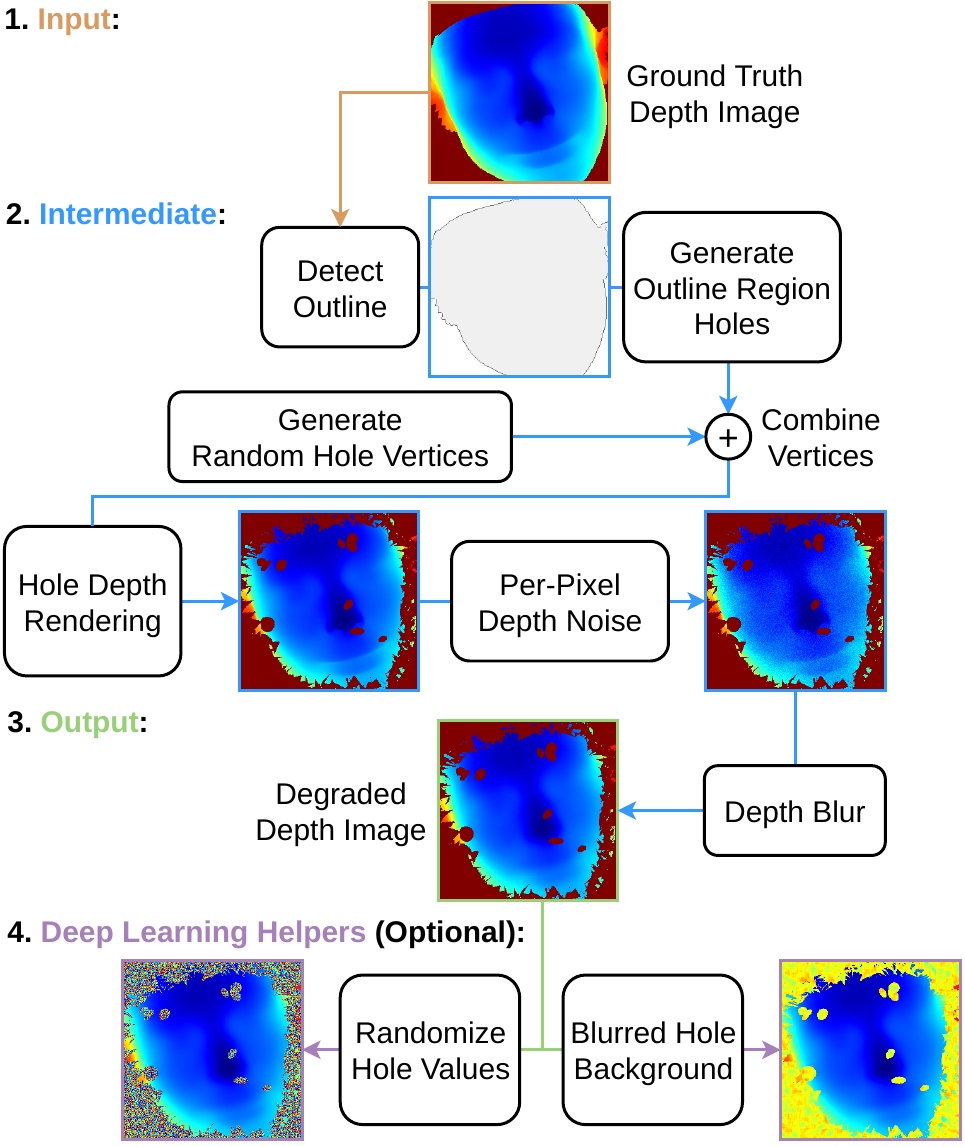}
	\caption{\label{fig:SynthDegProcess} The degradation synthesis process: Synthetic depth data is degraded in order to automatically generate a training set for the deep learning-based depth enhancer.
	}\vspace{-0.3cm}
\end{figure}

The second and final synthesis stage is the degradation process,
illustrated in \autoref{fig:SynthDegProcess}.
It uses the previously synthesized ground truth depth images as input,
``degrades'' them in a number of ways,
and then delivers those degraded depth images as output.
The image format - including the image dimensions (256×256) - remains identical to that of the ground truth input images throughout the process,
so that a direct comparison between degraded images and their corresponding ground truths is possible.

Three primary degradation types are synthesized:
Holes, noise, and blur.
Holes are invalid depth values and can be interpreted as the missing data in an image inpainting task.
For typical color image inpainting the hole synthesis should be computationally efficient and cover a diverse range of shapes that may occur in real use cases \citep{yuFreeFormImageInpainting2019}.
Random rectangular hole generation is considered insufficient \citep{18-InpaintingPartialConvolutions},
so newer approaches use random alteration of occlusion masks derived from two consecutive video frames \citep{18-InpaintingPartialConvolutions},
or more recently employ fully random generation based on varying lines \citep{yuFreeFormImageInpainting2019,shinPEPSIFastLightweight2020}, as well as cellular automata \citep{ntavelisAIM2020Challenge2020}.
The depth hole inpainting here is a simpler task insofar
that only depth face images are relevant,
that real holes are not defined by users,
and that there should be no strong and thus detrimental falsification of the original data.
So the enhancer can be less complex
and extreme inpainting for very large hole areas as in \citep{ntavelisAIM2020Challenge2020} is not required or sensible.
In comparison to typical color image inpainting,
using pairs of real ground truth and degraded images may be more reasonable for the face depth hole inpainting task,
but we do not attempt this due the issues previously discussed in \autoref{sec:datasets}.
Instead we also opted to use fully random hole generation.
This may however be beneficial,
since even a seemingly sufficient real dataset could miss scenarios or overrepresent hole areas,
which should not be the case with random holes for the same number of training samples.

The hole shapes are generated in the form of two different variants:
One variant places holes along the face outline,
which is detected by tracing the synthesized ground truth input image's non-hole depth pixels.
These holes ``erode'' the outline to some degree,
as this is likely to occur in reality,
see \eg{} \autoref{fig:ex-rs--01--2019-09-14T00-31-41}.
The second generator variant places a number of holes randomly across the image.
Individual hole parameters such as size and orientation are also randomized within a given range for both variants,
and the shapes of both variants are combined to render them in a single step on the GPU.

Next, random noise is added to (or subtracted from)
each remaining non-hole pixel's depth value,
to simulate the camera- and range-dependent universal depth uncertainty.
For the synthesis we choose a 15mm error,
which corresponds to real measured error values for the RealSense D435 in \citep{RealSenseDepthTuning} at 2.0m.
So according to the \citep{RealSenseDepthTuning} measurements this value is larger than necessary for the targeted sub-1m distance,
to compensate for any potential detrimental deviations in real recordings (\eg{} if another unknown camera type with worse error values were to be used).

At this point the intermediate form of the degraded image is still as sharp\slash crisp as the synthesized ground truth image,
while real consumer camera depth images as in \autoref{fig:CamsRsDistEx0} appear less clearly defined at a similar resolution.
The degradation synthesis approximates this lack of detail by blurring the non-hole depth pixels of the image,
which means that in a sub-process the depth values are averaged over a certain area per pixel.
With the end of this step the primary degradation synthesis is completed.

However, further degradation steps can be taken to alter the hole pixels' depth values,
thus altering the image's ``background''.
The hole pixel values can simply be fully randomized,
or randomized to a certain depth range and then blurred for a somewhat more predictable\slash plausible effect.
This intentionally unrealistic degradation step is meant to help the deep learning enhancer training
to better differentiate between important (\ie{} the face) and unimportant areas (\ie{} anything else in the background),
since real depth images may naturally contain various non-hole pixels in the non-face area,
representing \eg{} other recorded objects positioned in the same depth range as the face.
So this kind of degradation is not used for test data in any of the evaluations.
The idea is that the unaltered hole pixels may cause the deep learning enhancer
to expect the same kind of clearly identifiable hole\slash non-face pixels in real input (a case of overfitting),
whereas randomized hole values will force the network training to focus on more reliable features.
This focus on the face area stands in contrast to typical color image enhancement tasks, where the whole image is assumed to be relevant.

\begin{figure}[tb]
	\centering
	\includegraphics[width=0.8220\linewidth]{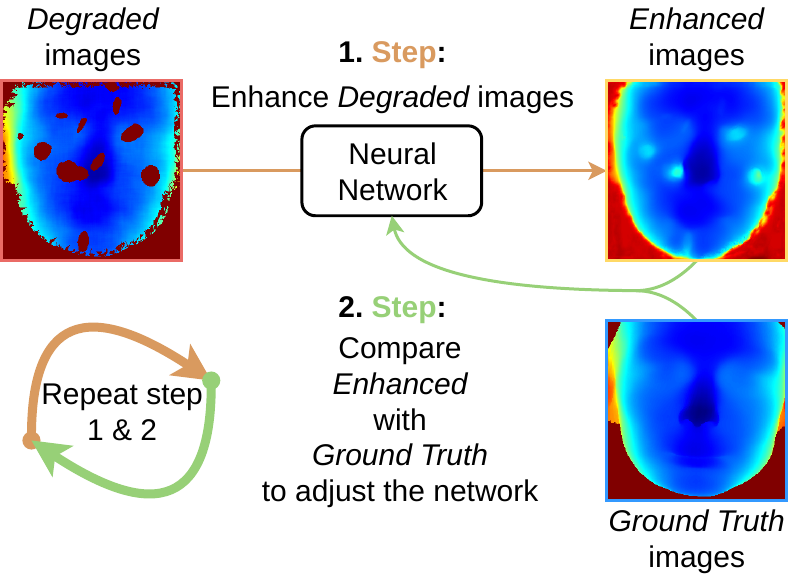}\vspace{-0.5cm}
	\caption{\label{fig:EnhancerDlConcept} Deep learning enhancer training concept visualization: Pairs of ground-truth depth images and corresponding degraded depth images are employed to train a neural network for depth enhancement.
	}\vspace{-0.3cm}
\end{figure}

\begin{figure*}[tb]
	\centering
	\includegraphics[width=0.7\linewidth]{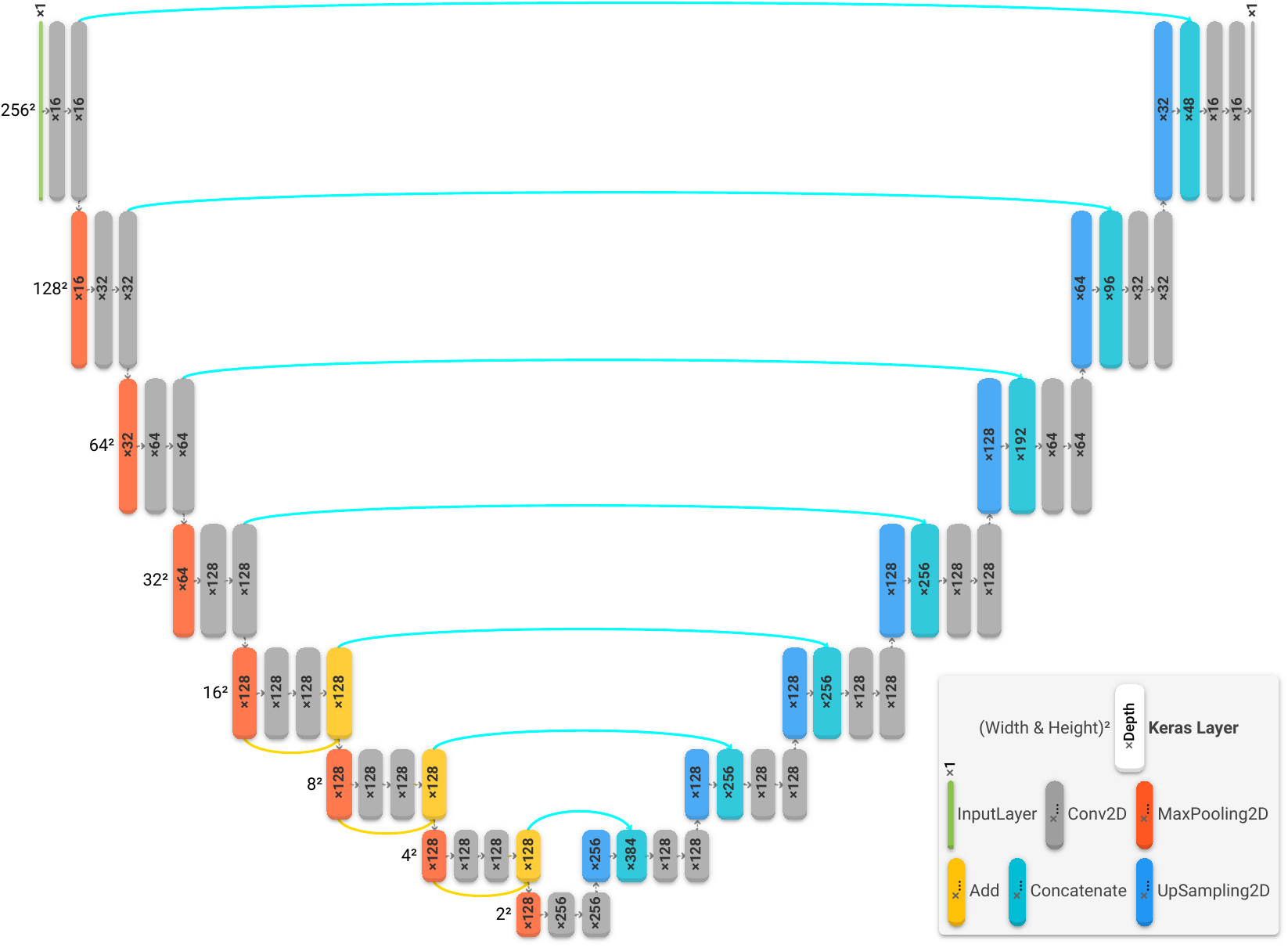}
	\caption{\label{fig:ArchDeltaVis} U-Net-like deep learning depth image enhancer architecture ``DL-Delta'' using Keras layers \citep{15-Keras}.}\vspace{-0.3cm}
\end{figure*}

\subsection{Deep Learning-based Depth Enhancer}

The deep learning depth enhancer network is created using Keras \citep{15-Keras}.
As illustrated in \autoref{fig:EnhancerDlConcept},
a network is trained using the synthesized ground truth and degraded images
as input.
Conceptually, the network first attempts to enhance a degraded input image,
then the ``enhanced'' image is compared against the ground truth image. With respect to the differences observed the network weights are adjusted.
By repeating this process with a preferably large amount of varied degraded\slash ground truth image pairs,
a network can be progressively trained to enhance an image.

Since the intended effect of the depth enhancement is to correct errors of the face depth representation,
\eg{} to fill holes,
and to generally increase the detail of the face depth image,
other, similar deep learning tasks with existing implementations such as
super-resolution \citep{19-DlSuperResolutionSurvey, 19-DlSuperResolutionWoGan},
image denoising \citep{18-DlDenoisingSurvey, 18-Noise2Noise},
or image inpainting \citep{18-InpaintingPartialConvolutions}
were considered for the depth enhancement system:

Among the cited prior works,
variants of the U-Net network architecture \citep{15-UNet} were used successfully
for super-resolution \citep{19-DlSuperResolutionWoGan}, denoising \citep{18-Noise2Noise} and inpainting \citep{18-InpaintingPartialConvolutions}.
Consequently a U-Net-like architecture was selected for the proposed deep learning depth enhancement as well.
Advancing the network architecture design is outside the scope of this work, since it is meant as a first feasibility study regarding the application of deep learning for face depth image enhancement
in the context of (PAD for) facial biometrics.
However,
simpler architectures which turn out to be insufficient for general color image enhancement
may be applicable to enhance face depth images:
Depth images require only one channel of information per pixel,
whereas typical color images have at least three (\eg{} red/green/blue).
The specialization on face depth images reduces the required complexity of the generational capabilities further.
And while typical color image enhancement may not have to care about security concerns,
here we do not want to overly falsify input by generating facial features where there should be none,
meaning that a network with less powerful generational capabilities might be preferable over one that produces good looking face images.

The name U-Net stems from the original U-shaped design of the network architecture proposed by Ronneberger \etal{} \citeAfterAuthors{15-UNet}.
\autoref{fig:ArchDeltaVis} depicts the smallest of our four similar U-Net-like network architectures,
which is also vaguely U-shaped.
The concept of an U-Net-like architecture can be summarized as follows:
First, the input is progressively scaled down (\ie{} the left side of the U-shape),
then the intermediate smallest representation is scaled up again (\ie{} the right U side),
whereby the downscaling stages' output is connected to
the corresponding upscaling stages.
Because each depth enhancer network takes one face depth image as input and outputs a presumably enhanced image in the same format,
\ie{} with the same resolution and a depth value per pixel,
the networks can be considered as a kind of (``denoising'') auto-encoder \citep{16-DeepLearning, 19-AutoEncoderWhatIsIt, 18-IntroductionToAutoEncoders}.
More information regarding the created and tested depth enhancer network configurations follows in \autoref{sec:enhancer-configurations}.

\section{Experiments}
\label{sec:experiments}

\newcommand{\captionCommonSynthetic}[1]{
Random synthetic sample #1.
Top row: The ground truth depth image and the degraded version, followed by hand-crafted enhancer output.
Bottom row: Output of DDRNet-dn and proposed deep learning-based enhancers.
The relevant face area is masked for illustrative clarity using the ground truth.
}
\newcommand{\captionCommonReal}{
Top row: The color image and the original depth image,
followed by hand-crafted enhancer output.
Bottom row: Output of DDRNet-dn and the proposed deep learning-based enhancers.
The relevant face area is masked for illustrative clarity using PRNet output for the color image.
}
\newcommand{\commonQuantitativeRealCaption}{Roughness and hole percentages for face regions in frontal KinectFaceDB \citep{14-KinectFaceDB} image variants and frontal RealSense D435 images.}

\newcommand{\commonCaptionFalsificationExample}[1]{
#1 KinectFaceDB enhancer falsification experiment samples.
The rows from top to bottom show images for the
\textit{Neutral}, \textit{OcclusionMouth}, and \textit{OcclusionPaper} variants respectively.
The relevant face area is masked as part of the evaluation based on
Dlib \citep{dlib09} facial landmark detector output
for the \textit{Neutral} color image.}

In this section the trained deep learning depth enhancers are compared against preexisting enhancement methods.
First, three preexisting hand-crafted enhancer types and the related work deep learning enhancer ``DDRNet-dn'' are explained.
Then the 15 experiment enhancer configurations are presented,
\ie{} the settings for the hand-crafted enhancers as well as the different trained deep learning enhancer networks.
These enhancer configurations are used in all of the remaining sections.

The difference between the experiments in this work and experiments from related depth enhancement works is that our experiments focus on enhancer properties specifically relevant to single face depth images in a biometric context.
For instance,
face depth PAD can be implemented \eg{} as a simple hand-crafted approach by examining depth values along facial landmarks.
The PAD may falsely reject depth input with real faces due to noise or holes,
so filling holes and smoothing noise can be sensible to avoid this.
Depth-based or depth-supported face recognition could likewise benefit from enhancement.
However, in contrast to unconstrained (or perceptual quality focused) depth enhancement found in related work,
depth enhancement for biometrics should avoid substantial falsification of the input data for the sake of security.
An enhancer should especially avoid the generation of face depth data where no face is present in the depth input,
since this could \ia{} circumvent the depth PAD even for simple presentation attack instruments such as printed face images.
As a result, the following experiments are concerned with enhancement in terms of depth hole and noise correction,
but also examine the degree of falsification.

We first present results in \autoref{sec:quantitative-synthetic} for the quantitative synthetic evaluation,
which compares the synthetic ground truth images to the enhancers' output for degraded variants.
Next, in \autoref{sec:quantitative-real}, we quantitatively assessed how roughness/noisiness and holes are affected by the enhancers for real data from the KinectFaceDB \citep{14-KinectFaceDB,14-KinectFaceDBSite}
and for in-house RealSense D435 data.
We then compared to what degree enhancers falsify occluded and unoccluded face depth image variants of the KinectFaceDB \citep{14-KinectFaceDB} in \autoref{sec:quantitative-falsification}.
The results in these evaluations show that the proposed deep learning enhancers are more generally effective than the preexisting enhancers.
In \autoref{sec:quantitative-landmarks}, the deep learning enhancers are further tested in a simple landmark-based PAD scenario,
again using images from the KinectFaceDB \citep{14-KinectFaceDB}.

In the last three subsections we provide qualitative examples
for enhancement of the synthetic images (\autoref{sec:samples-synthetic}),
the KinectFaceDB \citep{14-KinectFaceDB} images (\autoref{sec:samples-kinectfacedb}),
and for the custom RealSense D435 recordings produced as part of this work (\autoref{sec:samples-realsense}).

\subsection{Hand-crafted Enhancer Types}
\label{sec:handcrafted-enhancers}

The deep learning enhancers in this work are compared to three general hand-crafted enhancer types,
``decimation'', ``hole-filling'' and ``spatial'',
each of which is either equivalent to or a minor extension of a (post-processing) filter from the RealSense SDK \citep{RealSenseDepthPostProcessingGitHub}.
While the RealSense SDK as a whole is intended to be used with the eponymous RealSense cameras,
these filters can also be applied to depth images from other sources (\eg{} a camera from another company).
Here they are called
``hand-crafted'' because no machine learning is used,
and ``general'' because these filters\slash enhancer types are meant to apply to depth images in general, instead of being specialized to handle depth images of faces in particular.
All of these enhancer types have at least one parameter that controls their operations.
Thus, for the sake of clarity, it is probably sensible to explicitly call them ``enhancer types'',
while ``enhancers'' is separately referring only to specific parameter configurations of one single enhancer type.
Consecutively applying multiple enhancers is not considered in the experiments since this
would not be beneficial.

It should be noted that the RealSense SDK also provides some other filters that intentionally  are not used here:
The temporal filter \citep{RealSenseDepthPostProcessingGitHub} is not used since this work only considers individual depth images without additional information, \ie{} not temporally coherent video streams.
And the threshold filter \citep{RealSenseDepthPostProcessingGitHub} is not used because it only applies trivial (minimum\slash maximum value) depth thresholds to the input image,
which is redundant for this work's input images,
since these images already have their depth range normalized
to the face depth range.

\subsubsection{Decimation}

The decimation filter condenses the depth values from a window of input pixels to determine a single output pixel  \citep{18-RealSenseDepthPostProcessing,RealSenseDepthPostProcessingGitHub},
meaning that the output image produced by the filter is smaller then the input image.
However, this work's enhancer types are
required to output a depth image of equivalent size to their given input,
so that a direct quantitative comparison is possible.
The decimation enhancer type built on top of the decimation filter thus has to upscale the filter's output back to the original size,
introducing the
scaling
function's interpolation mode as an additional enhancer type parameter,
which is set to bilinear.
The filter itself has only one parameter, the ``magnitude'',
which specifies the filter's kernel size from 2×2 (the default) up to 8×8 pixels \citep{RealSenseDepthPostProcessingGitHub}.
For 2×2 and 3×3, one output pixel is the median depth value of the observed input pixels;
but for kernel sizes 4×4 and above, one output pixel is the mean depth value of the observed input pixels,
due to performance considerations \citep{RealSenseDepthPostProcessingGitHub}.
See \citep{RealSenseDepthPostProcessingGitHub} for a benchmark of all three filters,
which also shows how the decimation filter computation time drops when switching from magnitude 3 to magnitude 4.
Only non-hole input pixels
are taken into consideration during the decimation process \citep{RealSenseDepthPostProcessingGitHub},
so that the filter provides rudimentary hole-filling functionality \citep{RealSenseDepthPostProcessingGitHub}.

\subsubsection{Hole-filling}

The hole-filling filter \citep{RealSenseDepthPostProcessingGitHub}
is arguably the simplest among the three filters.
As the name suggests, the filter is meant exclusively to fill holes
(\ie{} by this work's definition pixels with the maximum depth value)
in the given depth input image.
It has only one parameter that specifies one of three operating modes where the center pixel represents a hole that is to be filled \citep{RealSenseDepthPostProcessingGitHub}:  ``Left'' refers to always choosing the value to the left, while ``Farest-From-Around'' and ``Nearest-From-Around'' choose the maximum and minimum non-hole depth value of chosen neighbor pixels, respectively.

The filter moves from the left to the right of each horizontal line of the image,
which means that all hole pixels will be filled,
except for cases where the initial neighboring pixel(s)
are holes.
All non-hole pixels remain unmodified.

\subsubsection{Spatial}

This filter's full name is ``spatial edge-preserving filter'' \citep{RealSenseDepthPostProcessingGitHub}.
It certainly is the functionally most complex of the hand-crafted methods used here,
and also the slowest according to the previously mentioned benchmark \citep{RealSenseDepthPostProcessingGitHub}.
The spatial filter primarily smooths non-hole depth values while preserving edges.
In addition, the filter has an optional built-in hole-filling mode.

There are four different parameters:
First, the number of iterations can be set.
Each iteration consists out of applying the filter horizontally (left to right, then right to left per line),
then vertically (top to bottom, then bottom to top),
which stands in contrast to the decimation \& hole-filling filters' behaviors,
both using only one pass to process an input image \citep{RealSenseDepthPostProcessingGitHub}.

Next, there are two parameters regarding the smoothing process itself:
``Smooth $\alpha$'' specifies the percentage used to determine how much of the currently processed pixel's value is retained in the output.
A value of 100\% means that no smoothing is applied, whereas 0\% (an invalid value) would be ``infinite'' smoothing \citep{RealSenseDepthPostProcessingGitHub}.
The other smoothing parameter, ``Smooth $\delta$'', defines the threshold that is used to recognize (and thus preserve) edges \citep{RealSenseDepthPostProcessingGitHub}.
More concretely, an edge is detected if the absolute difference between the currently processed pixel's value and the previous pixel's value is greater than $\delta$ \citep{RealSenseDepthPostProcessingGitHub}.

Lastly there is the ``Hole-filling (mode)'' parameter - not to be confused with the dedicated hole-filling filter.
This built-in hole-filling is optional and turned off by default.
To activate it, a radius of 2, 4, 8, 16 or infinite pixels can be chosen.
It operates by filling holes first left to right, then right to left per line.
If there are more hole pixels in succession than the selected radius,
the hole-filling stops until a non-hole pixel is encountered.
Holes are filled with the last non-hole pixel value read in the current line,
meaning they will not be filled if there are no preceding non-hole pixels in that line.
In comparison to the dedicated hole-filling filter,
this functionality is most similar to the ``Fill-From-Left'' mode,
with the differences being that the spatial filter's hole-filling is bidirectional
and has an optionally limited radius.

\begin{table}[htb]\vspace{-0.3cm}
	\caption{\label{tab:ex-enhancers} Hand-crafted enhancer configurations.}\vspace{-0.0cm}
	\centering
	\begin{tabular}{l c c}
	\toprule
		\textbf{Label} & \textbf{Type} & \textbf{Mode} \\
		\midrule
		De-M2 & Decimation  & Magnitude: 2 \\
		De-M8 & Decimation  & Magnitude: 8 \\
		Hf-FFA & Hole-filling & ``Farest-From-Around'' \\
		Hf-NFA & Hole-filling & ``Nearest-From-Around'' \\
		Hf-L & Hole-filling & ``Fill-From-Left'' \\
		Sp-Hf16 & Spatial & Hole-filling: 16 pixels \\
		Sp-HfU & Spatial & Hole-filling: $\infty$ (Unlimited) \\\bottomrule
	\end{tabular}\vspace{-0.1cm}
\end{table}

\begin{table}[htb]\vspace{-0.3cm}
	\caption{\label{tab:ex-enhancers-dl-tp} Proposed deep learning enhancer sizes.}\vspace{-0.0cm}
	\centering
	\begin{tabular}{l r}
	\toprule
		\textbf{Label} & \textbf{Trainable parameters} \\
		\midrule
		DL-Alpha & $6\,731\,041$ \\
		DL-Beta & $13\,746\,881$ \\
		DL-Gamma & $11\,593\,825$ \\
		DL-Delta (4 variants) & $4\,175\,713$ \\\bottomrule
	\end{tabular}
\end{table}

\subsection{DDRNet-dn Enhancer}
\label{sec:ddrnet-dn}

While there are no related works with models trained to enhance face depth images using only the depth image as input,
we include the depth-only ``denoising net'' part of DDRNet \citep{18-DDRNet} as ``DDRNet-dn'' in the experiments.
We use the official publicly available DDRNet GitHub \citep{19-DDRNet-GitHub} implementation with a pretrained model.
The DDRNet ``refinement net'' part is not used, since it requires color image input, and only depth input is permitted within our experiments.

Since DDRNet-dn conceptually is a denoising auto-encoder similar to our proposed deep learning enhancers,
the main difference is the considered application area and the training approach.
As the various following experiments show,
the given DDRNet-dn enhancement distorts the depth range, introduces artifacts, and does not effectively fill holes.
Except for the artifacts, this can likely be explained by the difference in training data, which consisted of (real) full body depth images instead of (synthetic) face-only depth images.
The ``checkerboard'' artifacts likely stem from the network architecture, as noted by \citep{18-DDRNet},
and could be limited to this particular publicly available implementation.

\subsection{Enhancer Configurations}
\label{sec:enhancer-configurations}
In all of the following experiment sections, 15 different enhancers are utilized, comprising seven proposed deep learning enhancers, seven preexisting hand-crafted enhancers,
and the DDRNet-dn deep learning enhancer from related work.
A 16th ``None'' variant shows results with no enhancement applied for comparison.

The hand-crafted enhancer configurations listed in \autoref{tab:ex-enhancers}
use the corresponding RealSense SDK default settings \citep{RealSenseDepthPostProcessingGitHub}
where not specified otherwise.
In addition to these we tested the ``Spatial'' enhancer type without hole-filling, which produced results comparable to no enhancement, so they were omitted.

\autoref{tab:ex-enhancers-dl-tp} provides an overview of the different architectures by size\slash complexity.
The evaluations of the proposed deep learning enhancer variations can be considered as an ablation study with respect to the network size and training approach:
DL-Alpha represents the first tested successful network architecture,
which was progressively trained with a variety of synthetic data, as will be described shortly.
DL-Beta is architecturally similar to DL-Alpha, but with a markedly increased filter size for the 2D convolution layers.
DL-Gamma's architecture is identical to DL-Alpha's,
except that the 2D convolution window kernel size
has been increased from 3×3 (used by the other networks) to 5×5,
thus again increasing the network complexity in comparison to DL-Alpha.
In contrast, the DL-Delta architecture depicted by \autoref{fig:ArchDeltaVis} is a shrunken and simplified version of DL-Alpha,
which means that the network requires less computational resources,
thus also allowing training in a shorter amount of time.

DL-Alpha, DL-Beta and DL-Gamma
were trained using $240\,000$ synthetic ground truth\slash degraded image pairs,
with $30\,000$ thereof being used in training for 10 epochs before moving on to the next $30\,000$.
All network models were saved only if their loss value improved during training.
The $240\,000$ degraded images employ the deep-learning-specific hole value randomization degradation.

Prior to this ``$240\,000$ pair training'', DL-Alpha was also trained
on  $70\,000$ synthesized image pairs without deep-learning-specific degradation,
and on $60\,000$ pairs using the deep-learning-specific blurred hole background degradation afterwards.
It is however plausible to assume that the latest ``$240\,000$ pair training'' predominantly determines the network's learned behavior, mostly overriding any behavior stemming from the prior training sessions, since their degraded image types were no longer represented in the training input.
DL-Beta and DL-Gamma were trained exclusively with the $240\,000$ set.

DL-Delta training used $60\,000$ ground truth images
that represent a subset of the aforementioned $240\,000$ training samples.
For each ground truth image three degraded variants are used for training:
One without deep-learning-specific degradation (\ie{} unmodified hole pixels), one with hole value randomization and one with blurred hole background degradation.
The three DL-Delta degradation input types are interleaved during training, 
so that $30\,000$ degraded images corresponding to $10\,000$ ground truths
are trained for 10 epochs each, thus taking 6 training iterations thereof to process all pairings with the $60\,000$ ground truth images.

DL-Delta-1 refers to the DL-Delta version which was trained in this manner once.
DL-Delta-2, DL-Delta-3 and DL-Delta-4 respectively repeated this training process based on their preceding network model (\eg{} DL-Delta-4 was thus trained four times in the described way).

\begin{figure}[htb]
	\centering
	\includegraphics[width=\linewidth]{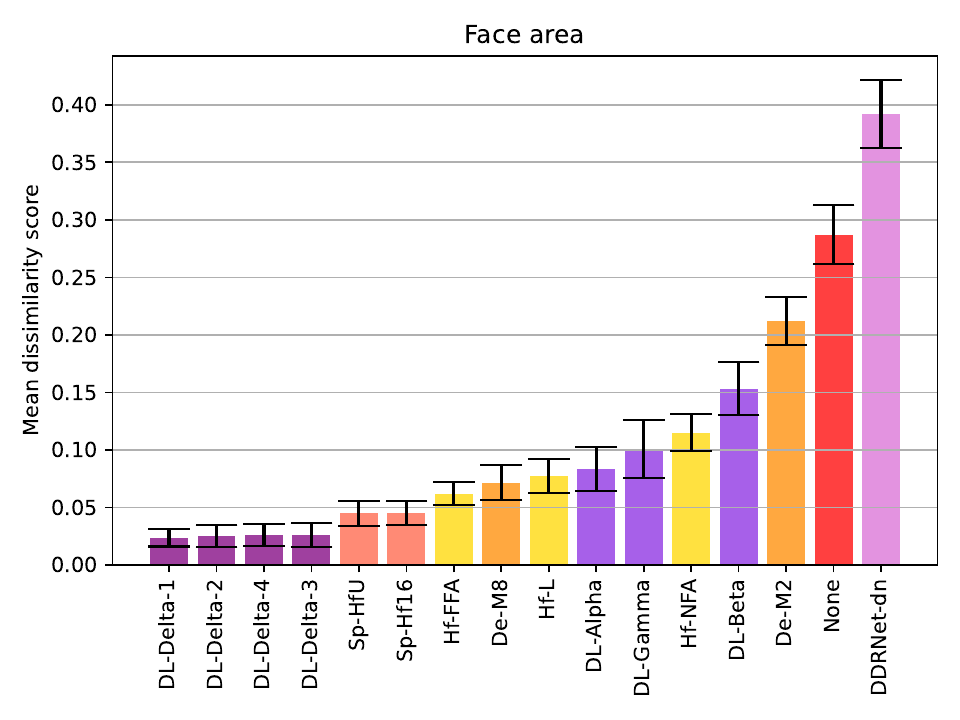}
	\vspace{-0.6cm}
	\caption{\label{fig:ex-se--RMSE}
	Mean RMSE results with standard deviation error bars.
	The proposed DL-Delta enhancers outperform the other enhancer types.
    Corresponds to \autoref{tab:ex-se--RMSE}.
	}
\end{figure}

\begin{table}
    \centering
    \caption{\label{tab:ex-se--RMSE}
    RMSE results as in \autoref{fig:ex-se--RMSE} sorted by mean $\bar{x}$, with standard deviation $s$, skewness and excess kurtosis.
    }
\begin{tabular}{r|cccc}
\toprule
\textbf{Enhancer} & \textbf{$\bar{x}$} & \textbf{$s$} & \textbf{Skewness} & \textbf{Kurtosis} \\
\midrule
DL-Delta-1 &  0.0238 &  0.0078 &  1.8831 &  8.5198 \\
DL-Delta-2 &  0.0249 &  0.0095 &  2.2523 & 11.4975 \\
DL-Delta-4 &  0.0260 &  0.0094 &  1.9481 &  8.7021 \\
DL-Delta-3 &  0.0261 &  0.0101 &  2.1560 & 10.6805 \\
    Sp-HfU &  0.0448 &  0.0107 & -0.0196 & -0.2635 \\
   Sp-Hf16 &  0.0449 &  0.0104 &  0.0004 &  0.0280 \\
    Hf-FFA &  0.0621 &  0.0098 &  0.9596 &  4.3058 \\
     De-M8 &  0.0716 &  0.0154 &  0.0382 &  0.1172 \\
      Hf-L &  0.0774 &  0.0151 &  1.2580 &  3.4937 \\
  DL-Alpha &  0.0836 &  0.0189 &  0.8079 &  5.9283 \\
  DL-Gamma &  0.1008 &  0.0254 &  0.2126 & -0.1884 \\
    Hf-NFA &  0.1151 &  0.0164 &  0.8171 &  0.9016 \\
   DL-Beta &  0.1532 &  0.0230 &  0.1246 &  0.1763 \\
     De-M2 &  0.2120 &  0.0211 &  0.2463 &  0.3012 \\
      None &  0.2872 &  0.0256 &  0.4468 &  0.5183 \\
 DDRNet-dn &  0.3918 &  0.0296 &  0.0129 &  0.3807 \\
\bottomrule
\end{tabular}
\end{table}

\subsection{Quantitative Synthetic Evaluation}
\label{sec:quantitative-synthetic}

\begin{figure*}[!htb]
	\centering
	\textbf{KinectFaceDB}\\
	\includegraphics[width=0.45\linewidth]{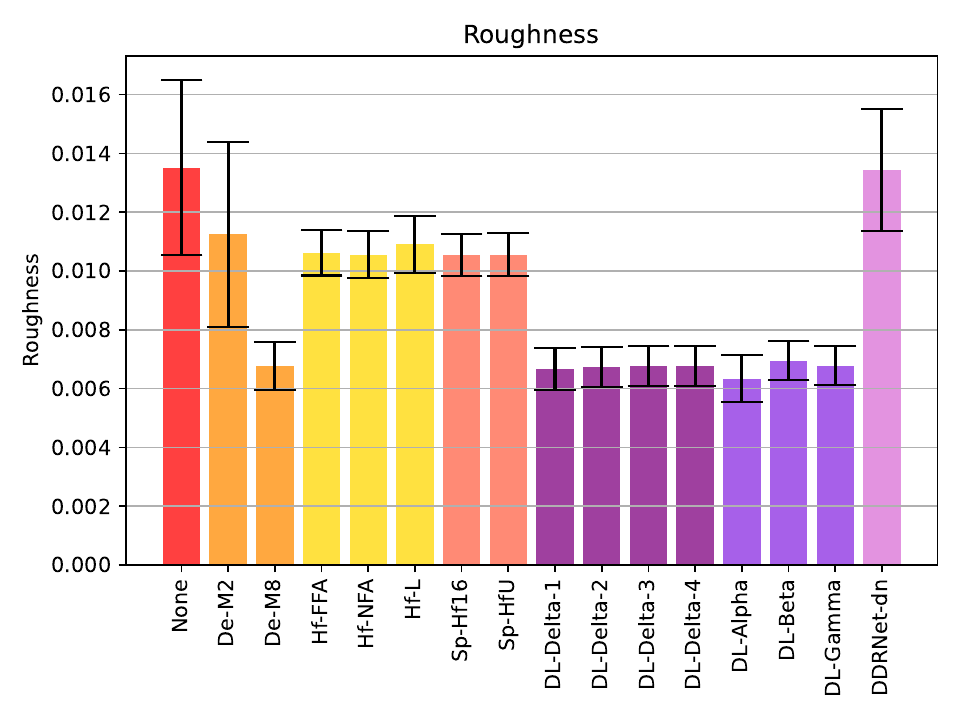}
	\includegraphics[width=0.45\linewidth]{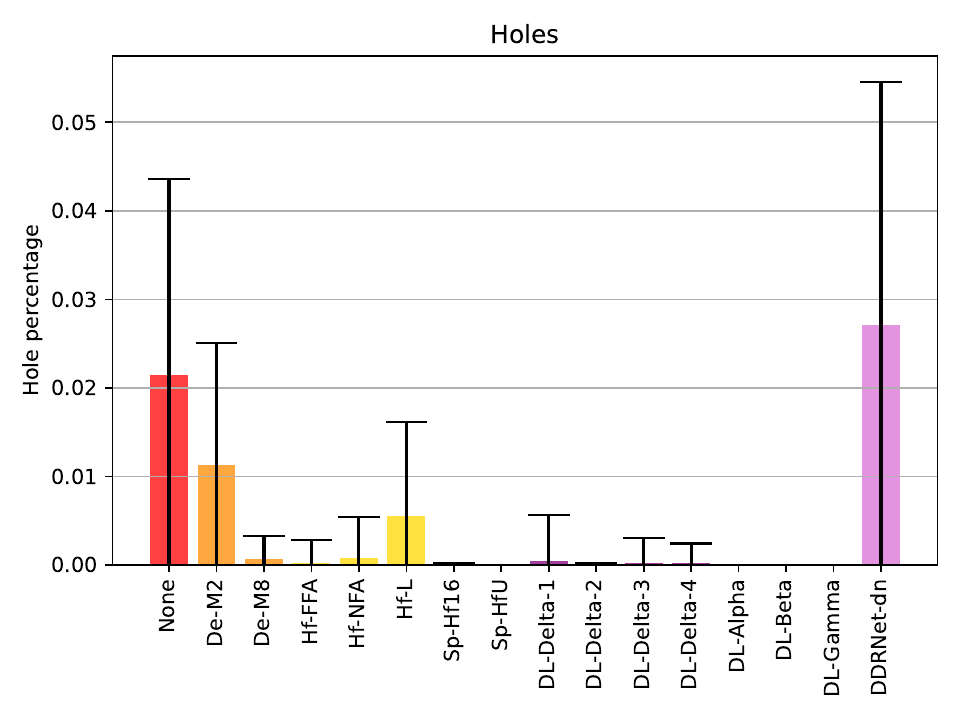}\\
	\textbf{RealSense D435}\\
	\includegraphics[width=0.45\linewidth]{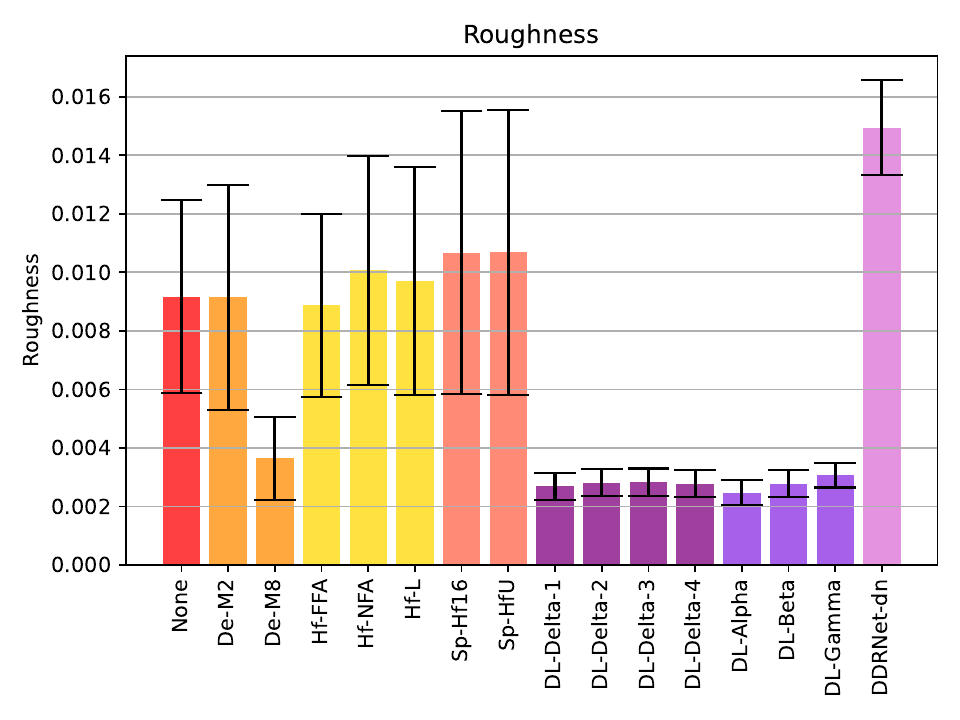}
	\includegraphics[width=0.45\linewidth]{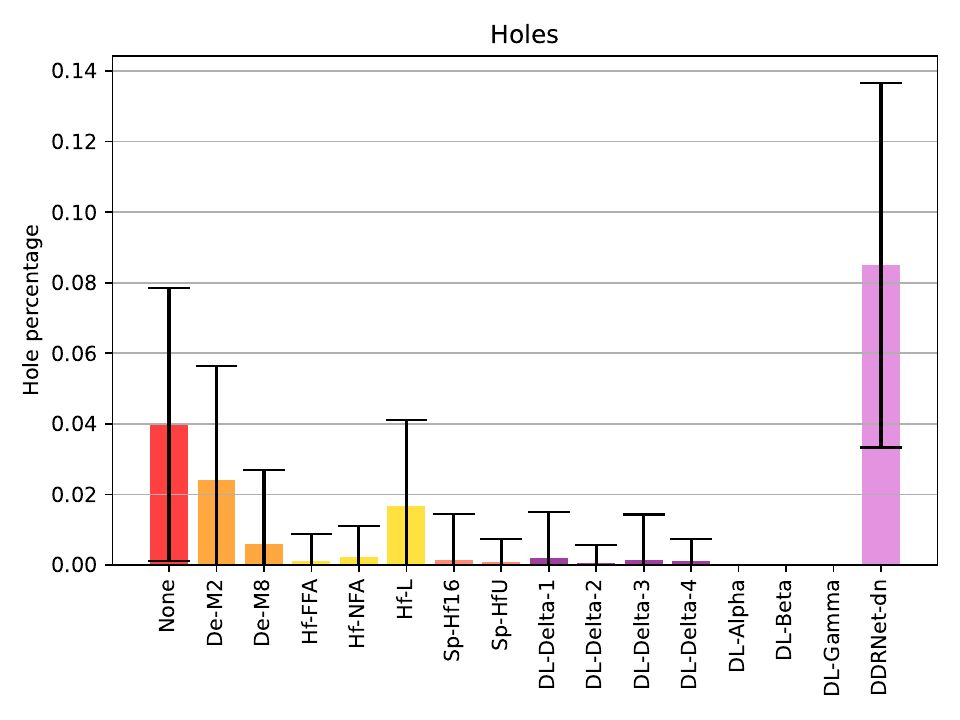}\\
    \caption{\label{fig:quantitative-real}\commonQuantitativeRealCaption{} Corresponds to \autoref{tab:quantitative-real}.}
    \vspace{-1em}
\end{figure*}

\newcommand{\syntheticPairCount}{$100\,000$}

\syntheticPairCount{} synthesized pairs of ground truth and degraded depth images have been used in this evaluation.
Source data for the synthesis has been randomly selected from the MF2 distractor subset, with data already used during the
deep learning enhancer training being excluded from the candidate list.
Identical ranking and approximately equivalent individual result values were also obtained with only $10\,000$ pairs, so we assume that additional synthetic data of the same kind would not alter this evaluation.
Since it is possible to \ia{} change the degradation synthesis parameters to multiply the amount of available evaluation pairs,
testing all feasibly available data is unfeasible due to
time
constraints.

The \syntheticPairCount{} degraded images are provided as input to the experiment enhancers,
and each enhancer's output image is compared to the associated ground truth image.
Visual examples for these synthesized pairs and the corresponding enhancer results
are showcased in the next subsection,
while this section presents the quantitative numerical results for all \syntheticPairCount{} test pairs.
The dissimilarity measure RMSE (``Root Mean Squared Error'') is used, $RMSE=\sqrt{\frac{\sum_{i=1}^{n}{(p_{gt_i}-p_{e_i})^2}}{n}}$.

In the RMSE equation,
$p_{gt_i}$ refers to the ground truth pixels,
$p_{e_i}$ to the enhancer output pixels,
and $n$ to the number of compared pixel pairs.
All paired pixels share identical image coordinates, with all images being 256×256.
The pixel indices are denoted by $i$. The RMSE results are presented in \autoref{fig:ex-se--RMSE}
and \autoref{tab:ex-se--RMSE}.
Results are
computed for the face image area,
which is defined by the non-hole pixels in the ground truth image ($n$ then being equivalent to the count thereof).

The DL-Delta enhancer versions dominate the results
with the lowest error values. Similarly, over the full image area ($n=256^2$) all deep learning enhancers showed better performance than the hand-crafted enhancers,
but detailed results are not included since only the enhancement in the face area is relevant.
Despite their higher network complexity,
the proposed non-Delta deep learning variants 
-- DL-Alpha, DL-Beta and DL-Gamma --
do not surpass the best performing hand-crafted enhancers when only the important face area is considered.
The likely explanation for the non-Delta deep learning variant inferiority is
that these were predominantly trained on data with deep-learning-specific hole degradation applied,
standing in contrast to the DL-Delta versions,
which were trained with a mix of images with both kinds of deep-learning-specific hole degradation as well as regular holes.
Among the DL-Delta versions, DL-Delta-1 appears universally superior,
albeit only by a small margin.
This may be explained as a mild case of overfitting probably experienced by the subsequent training iterations on the same data to create DL-Delta-2, DL-Delta-3 and DL-Delta-4,
with DL-Delta-1 seemingly utilizing the network architecture to approximately full effect already (possible further training with different data notwithstanding).

Of the hand-crafted enhancers, the Sp-Hf16 and Sp-HfU spatial enhancers with hole-filling capabilities delivered the best results in the face area.
It is important to keep in mind that these results are based on only approximately realistic synthesized data
degradation with some samples shown in the next subsection. That is, the real camera samples
and supporting quantitative experiments on real data
following in the remaining subsections should be taken into consideration as well.

The DDRNet-dn enhancer led to worse results than no enhancement.
This can be explained by the introduced checkerboard artifacts and the depth range falsification issue,
which is more clearly shown by the following evaluations.
As already stated in \autoref{sec:ddrnet-dn},
the likely most important difference between DDRNet-dn and the proposed deep learning enhancers is the training approach.
While parts of the DDRNet-dn training images depicted faces,
it was not explicitly trained for face depth enhancement,
so the ineffectiveness here may not be surprising.

\subsection{Quantitative Real Data Evaluation}
\label{sec:quantitative-real}

\begin{table*}[!htb]
    \centering
    \caption{\label{tab:quantitative-real}
        \commonQuantitativeRealCaption{}
        Numbers outside brackets show the mean, numbers inside brackets the sample standard deviation.
        All values are multiplied by 100.
        Corresponds to \autoref{fig:quantitative-real}.
    }
\begin{tabular}{r|cc|cc}
\toprule
\textbf{Enhancer} & \multicolumn{2}{c}{\textbf{KinectFaceDB}} & \multicolumn{2}{c}{\textbf{RealSense D435}} \\
 & \textbf{Roughness} & \textbf{Holes} & \textbf{Roughness} & \textbf{Holes} \\
\midrule
      None & 1.35 (0.30) & 2.15 (2.21) & 0.92 (0.33) & 3.98 (3.88) \\
     De-M2 & 1.12 (0.31) & 1.13 (1.38) & 0.91 (0.38) & 2.40 (3.22) \\
     De-M8 & 0.68 (0.08) & 0.06 (0.26) & 0.36 (0.14) & 0.60 (2.08) \\
    Hf-FFA & 1.06 (0.08) & 0.03 (0.26) & 0.89 (0.31) & 0.12 (0.74) \\
    Hf-NFA & 1.06 (0.08) & 0.08 (0.45) & 1.01 (0.39) & 0.23 (0.89) \\
      Hf-L & 1.09 (0.10) & 0.55 (1.06) & 0.97 (0.39) & 1.68 (2.43) \\
   Sp-Hf16 & 1.05 (0.07) & 0.00 (0.02) & 1.07 (0.48) & 0.14 (1.30) \\
    Sp-HfU & 1.06 (0.07) & 0.00 (0.00) & 1.07 (0.49) & 0.07 (0.67) \\
DL-Delta-1 & 0.67 (0.07) & 0.04 (0.52) & 0.27 (0.05) & 0.21 (1.31) \\
DL-Delta-2 & 0.67 (0.07) & 0.00 (0.02) & 0.28 (0.05) & 0.06 (0.52) \\
DL-Delta-3 & 0.68 (0.07) & 0.02 (0.29) & 0.28 (0.05) & 0.15 (1.28) \\
DL-Delta-4 & 0.68 (0.07) & 0.02 (0.22) & 0.28 (0.05) & 0.11 (0.62) \\
  DL-Alpha & 0.63 (0.08) & 0.00 (0.00) & 0.25 (0.04) & 0.00 (0.00) \\
   DL-Beta & 0.69 (0.07) & 0.00 (0.00) & 0.28 (0.05) & 0.00 (0.00) \\
  DL-Gamma & 0.68 (0.07) & 0.00 (0.00) & 0.31 (0.04) & 0.00 (0.00) \\
 DDRNet-dn & 1.34 (0.21) & 2.71 (2.74) & 1.49 (0.16) & 8.50 (5.17) \\
\bottomrule
\end{tabular}
\end{table*}

This experiment assesses enhancer performance in terms of the roughness and hole percentage in the face region of real images from the KinectFaceDB \citep{14-KinectFaceDB}
and from an in-house RealSense D435 dataset.
The roughness (or noisiness) is computed per pixel using a simple kernel that incorporates the 8 neighboring pixels:
\[
\begin{pmatrix}
-1/8 & -1/8 & -1/8 \\
-1/8 & 1 & -1/8 \\
-1/8 & -1/8 & -1/8 \\
\end{pmatrix}
\]

To select the face region,
binary masks are created based on Dlib \citep{dlib09} facial landmark detector output
for the \textit{Neutral} color image variants of the KinectFaceDB,
and for color images for the RealSense D435 dataset.
For the KinectFaceDB
\textit{Neutral} variants are used to mask the other frontal variants because the landmark detector output for various images of the other variants is incorrect, which is especially true for the occlusion variants (\textit{OcclusionEyes}/\textit{OcclusionMouth}/\textit{OcclusionPaper}).
It is possible to use the \textit{Neutral} masks to circumvent this problem for other frontal variants, since the faces are mostly centered,
but the \textit{LeftProfile}/\textit{RightProfile} variants were excluded from this experiment.
For the RealSense D435 dataset there are no occlusion variants and only frontal images were selected.

\autoref{fig:quantitative-real} and \autoref{tab:quantitative-real} show the results of this experiment as the mean and sample standard deviation over the face areas of all examined images.
We only show the combined results here because the relative differences between the enhancer results are similar if the variants are examined in isolation.
The proposed deep learning enhancers decrease the roughness/noisiness to the greatest degree among the enhancers,
and fill holes.
Some of the hand-crafted enhancers naturally fill all holes as well,
but are more limited when it comes to simultaneously handling noise,
as indicated by these results in conjunction with our synthetic experiment and qualitative examination.
These other results also show that the proposed deep learning enhancer output is not smoothed too much for depth PAD,
so the lowest roughness results here are likely above a functionally acceptable minimum for face depth PAD in general.
Similar to the synthetic quantitative evaluation, the DDRNet-dn enhancer was not effective for roughness and hole reduction.

\subsection{Falsification Evaluation}
\label{sec:quantitative-falsification}

\begin{figure*}[htb]
	\centering
	\includegraphics[width=\linewidth]{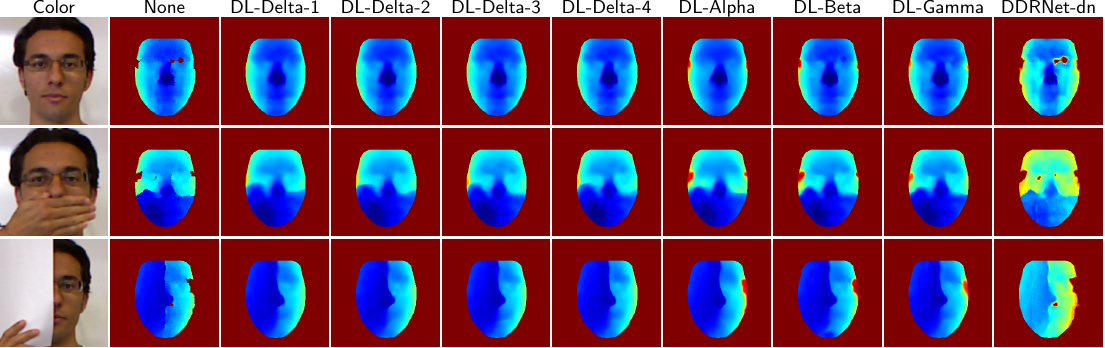}
    \caption{\label{fig:falsification-example}
    \commonCaptionFalsificationExample{0025\_s1}
    }
\end{figure*}

\begin{table}[htb]
    \centering
    \caption{Quantitative results for the input falsification experiment, with RMSE values in percent.
    Numbers outside brackets show the mean, numbers inside brackets the sample standard deviation.
    \textit{None}, \textit{Mouth}, \textit{Paper} occlusion refers to the \textit{Neutral}, \textit{OcclusionMouth}, \textit{OcclusionPaper} KinectFaceDB \citep{14-KinectFaceDB} variants respectively.
    }
\begin{tabular}{r|ccc}
\toprule
\textbf{Enhancer} & \textbf{None} & \textbf{Mouth} & \textbf{Paper} \\
\midrule
DL-Delta-1 & 1.51 (0.53) & 4.08 (4.80) & 2.90 (3.91) \\
DL-Delta-2 & 1.50 (0.59) & 3.78 (4.16) & 2.77 (3.55) \\
DL-Delta-3 & 1.51 (0.57) & 3.52 (3.86) & 2.93 (3.55) \\
DL-Delta-4 & 1.54 (0.58) & 3.87 (4.52) & 2.99 (3.74) \\
  DL-Alpha & 1.89 (0.67) & 2.73 (0.88) & 2.98 (1.42) \\
   DL-Beta & 2.01 (0.72) & 2.51 (0.82) & 2.48 (1.11) \\
  DL-Gamma & 1.65 (0.46) & 2.28 (0.61) & 2.32 (0.94) \\
 DDRNet-dn & 12.51 (5.13) & 17.02 (5.29) & 13.31 (6.14) \\
\bottomrule
\end{tabular}
    \label{tab:falsification}
\end{table}

While the other experiments are concerned to answer how well the enhancers fix degradations in the face depth input images,
this experiment aims to answer whether the deep learning enhancers hallucinate face depth into non-face depth input.
Avoiding this is essential for depth PAD - otherwise the ``enhancement'' may help attackers circumvent the PAD.
To assess this,
we quantitatively evaluated the degree of falsification in terms of RMSE between input and enhanced depth for unoccluded and occluded image variants in the KinectFaceDB \citep{14-KinectFaceDB}.
The preprocessing and masking is identical to that described in \autoref{sec:quantitative-real},
meaning that we only evaluate the RMSE within the face area defined
using Dlib \citep{dlib09} facial landmark detector output for the \textit{Neutral} variant.
In addition to ignoring the background, we also ignore the hole area of the unenhanced depth input, to focus on the falsification of existing non-hole depth pixels.

We use \textit{Neutral} as the unoccluded baseline variant for comparison against \textit{OcclusionMouth} and \textit{OcclusionPaper}.
These two occlusion variants comprise images that include the lower half of the face with hand,
or one side of the face with a piece of paper, respectively.
The presence of a partial face in each occluded image should make it easier to confuse a vulnerable deep learning enhancer, so robustness against these images also implies robustness against trivial attacks consisting only of \eg{} a bent paper with a printed face image.

\autoref{tab:falsification} shows the quantitative results.
For our deep learning enhancers,
there is no substantial increase in RMSE values and thus falsification between the unoccluded variant and the two occluded variants,
which entails that no faces appear to be hallucinated into the occluded halves of the face areas.
The DDRNet-dn enhancer likewise does not hallucinate faces into non-face areas, and its results comprise fewer changes to the holes, but the introduced artifacts and depth range falsification leads to distinctly worse results for this experiment.
While it would be possible to partially mitigate the depth range issue via depth adjustments with separately derived parameters,
this would constitute an unfair advantage over the other enhancer types that only use the depth input image.
Manual examination of the output supports the conclusions,
with one example being \autoref{fig:falsification-example}.

\subsection{Landmark Evaluation}
\label{sec:quantitative-landmarks}

\begin{figure}[!htb]
    \centering
    \includegraphics[width=0.9\linewidth]{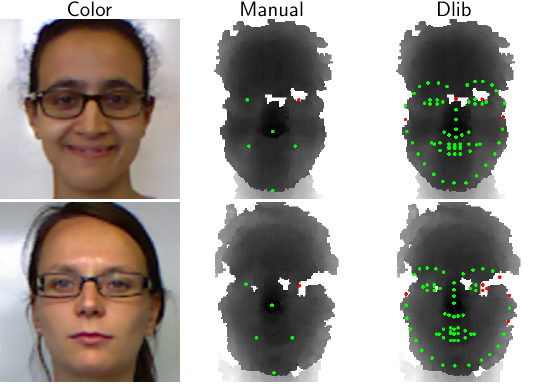}
    \caption{\label{fig:example-landmarks} Landmark depth error examples.}
\end{figure}

\begin{figure}[htb]
	\centering
	\includegraphics[width=0.8\linewidth]{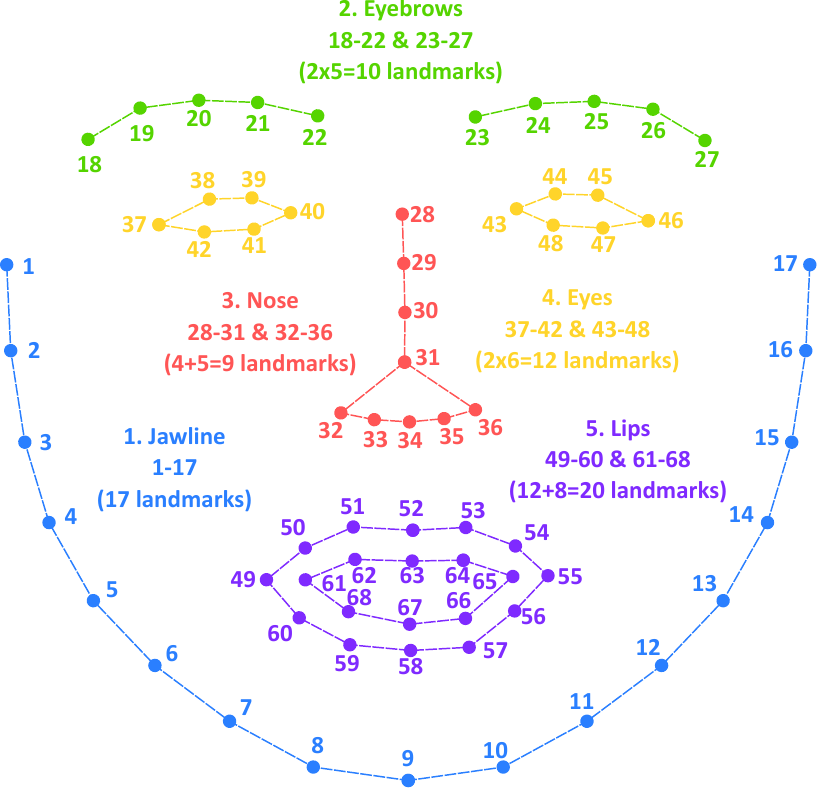}
    \caption{\label{fig:pad-landmarks}68 Dlib \citep{dlib09} detector landmarks.}
    \vspace{-0.2cm}
\end{figure}

\begin{table}[!htb]
    \centering
    \caption{\label{tab:quantitative-landmarks}
    Landmark evaluation results.
    Lower is better, results range from 0 to 1.
    ``Std. dev./Min./Max.'' is the sample standard deviation/minimum/maximum, respectively.
    }
\begin{tabular}{r|cccc}
\toprule
\multicolumn{5}{c}{\textbf{KinectFaceDB manual landmark holes}} \\
\textbf{Enhancer} & \textbf{Mean} & \textbf{Std. dev.} & \textbf{Min.} & \textbf{Max.} \\
\midrule
      None &      0.236 &      0.083 &      0.167 &      0.333\\
DL-Delta-1 &      0.000 &      0.000 &      0.000 &      0.000\\
DL-Delta-2 &      0.000 &      0.000 &      0.000 &      0.000\\
DL-Delta-3 &      0.000 &      0.000 &      0.000 &      0.000\\
DL-Delta-4 &      0.000 &      0.000 &      0.000 &      0.000\\
  DL-Alpha &      0.000 &      0.000 &      0.000 &      0.000\\
   DL-Beta &      0.000 &      0.000 &      0.000 &      0.000\\
  DL-Gamma &      0.000 &      0.000 &      0.000 &      0.000\\
 DDRNet-dn &      0.162 &      0.027 &      0.000 &      0.167\\
\bottomrule
\end{tabular}
\begin{tabular}{r|cccc}
\multicolumn{5}{c}{\textbf{Dlib landmark holes}} \\
\textbf{Enhancer} & \textbf{Mean} & \textbf{Std. dev.} & \textbf{Min.} & \textbf{Max.} \\
\midrule
      None &      0.042 &      0.025 &      0.015 &      0.162\\
DL-Delta-1 &      0.000 &      0.001 &      0.000 &      0.015\\
DL-Delta-2 &      0.000 &      0.000 &      0.000 &      0.000\\
DL-Delta-3 &      0.000 &      0.001 &      0.000 &      0.015\\
DL-Delta-4 &      0.000 &      0.001 &      0.000 &      0.015\\
  DL-Alpha &      0.000 &      0.000 &      0.000 &      0.000\\
   DL-Beta &      0.000 &      0.000 &      0.000 &      0.000\\
  DL-Gamma &      0.000 &      0.000 &      0.000 &      0.000\\
 DDRNet-dn &      0.040 &      0.027 &      0.000 &      0.147\\
\bottomrule
\end{tabular}
\begin{tabular}{r|cccc}
\multicolumn{5}{c}{\textbf{Dlib-landmark-based geometric PAD error}} \\
\textbf{Enhancer} & \textbf{Mean} & \textbf{Std. dev.} & \textbf{Min.} & \textbf{Max.} \\
\midrule
      None &      0.404 &      0.441 &      0.015 &      1.000\\
DL-Delta-1 &      0.054 &      0.055 &      0.000 &      0.721\\
DL-Delta-2 &      0.055 &      0.053 &      0.000 &      0.691\\
DL-Delta-3 &      0.053 &      0.053 &      0.000 &      0.706\\
DL-Delta-4 &      0.054 &      0.057 &      0.000 &      0.765\\
  DL-Alpha &      0.046 &      0.052 &      0.000 &      0.691\\
   DL-Beta &      0.053 &      0.052 &      0.000 &      0.691\\
  DL-Gamma &      0.056 &      0.056 &      0.000 &      0.765\\
 DDRNet-dn &      0.101 &      0.048 &      0.029 &      0.618\\
\bottomrule
\end{tabular}
\vspace{-2em}
\end{table}

While the prior evaluations show the relative effectiveness of the proposed deep learning enhancers,
we further tested whether they close holes at actual facial landmark positions with values suitable for PAD.
\autoref{fig:example-landmarks} shows examples for depth images with landmark hole errors,
\autoref{fig:pad-landmarks} illustrates the Dlib landmark structure,
and \autoref{tab:quantitative-landmarks} lists the evaluation results.
All proposed deep learning enhancers did close almost all holes at facial landmark positions,
both for the six landmarks manually defined in KinectFaceDB and for the 68 landmarks defined by Dlib's \citep{dlib09} facial landmark detector.
The evaluations were carried out on the subset of the KinectFaceDB \citep{14-KinectFaceDB} depth images with at least one landmark hole depth error,
which amounted to 77 images for the manual landmarks and 245 images for the Dlib landmarks.
Only non-``Occlusion'' variants were considered for the Dlib landmark test,
since the landmark detector relies on the corresponding color image, which often results in flawed output for occluded variants.

The Dlib landmarks were further used as input for a simple geometric frontal PAD function.
This function declared each landmark as either valid or invalid,
and computed the PAD error as the ratio of invalid landmarks to the total landmark count (\ie{} 68).
Landmarks were declared as invalid if they had a hole depth value,
or if their depth was nearer than the tip of the nose landmark depth,
or, for the jawline landmarks, if their depth decreased from the frontal chin landmark center (landmark 9 in \autoref{fig:pad-landmarks}) relative to the preceding landmark (\eg{} landmark 7 was marked invalid if landmark 8 had a greater depth value).
The proposed deep learning enhancers did effectively reduce the PAD error,
albeit not to zero, in contrast to the near-complete landmark depth hole reduction (see \autoref{tab:quantitative-landmarks}).

As to be expected from the prior evaluations,
DDRNet-dn did not achieve equivalent results,
but was not completely ineffective either.

We conducted an additional small-scale qualitative examination of the DL-Delta-1 enhancer for a real-time depth PAD test application using the RealSense D435.
A hand-crafted geometric depth PAD function similar to the quantitative evaluation variant compared depth values at Dlib landmarks,
with sufficiently large or implausible depth deviations leading to rejections of the input.
PAD decisions were temporal in the sense that per-frame decisions were observed over time to form a final decision for the detected face,
to increase the robustness.
The simple depth-only PAD function proved sufficient to detect low-effort attacks such as printed out faces (including bent paper).
DL-Delta-1 enhancement aided the PAD method's stability,
confirming that it is viable for PAD use.

\subsection{Synthetic Samples}
\label{sec:samples-synthetic}

\begin{figure*}[htb]
	\centering
	\includegraphics[width=\linewidth]{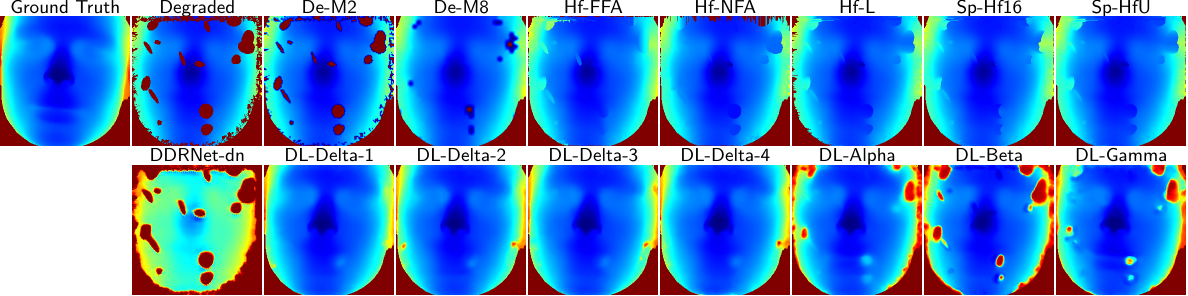}\vspace{-0.2cm}
	\caption{\label{fig:ex-synth-samples--2}
	\captionCommonSynthetic{1}
	}\vspace{-0.0cm}

\vspace*{\floatsep}

	\centering
	\includegraphics[width=\linewidth]{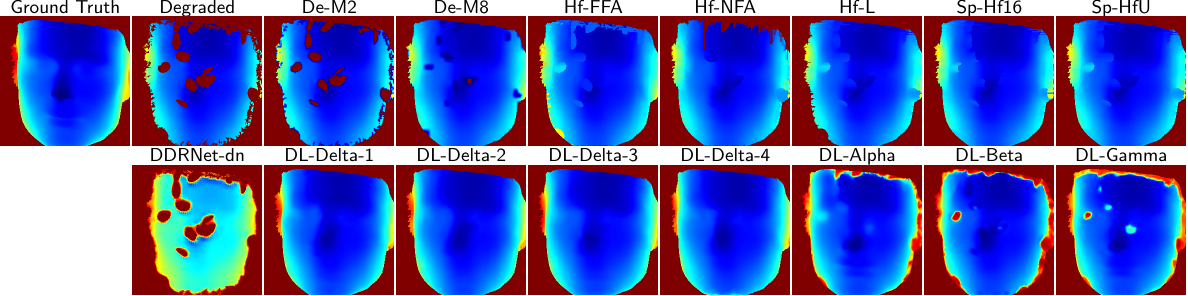}\vspace{-0.2cm}
	\caption{\label{fig:ex-synth-samples--4}
	\captionCommonSynthetic{2}
	}\vspace{-0.4cm}
\end{figure*}

\autoref{fig:ex-synth-samples--2} and \autoref{fig:ex-synth-samples--4} show randomly selected synthetic sample images together with their corresponding enhancer output,
as used in the quantitative synthetic evaluation.
All of the depicted depth images have been colorized to improve the visibility of depth differences.

The top left of each figure shows the synthetic ground truth,
to which the enhancer output was compared against in the RMSE evaluation,
and the degraded image,
which was provided as input to the enhancers.

Findings of the numeric evaluation results are reflected here:
The similar output of the four DL-Delta versions shows the enhancers' superior hole-correction capabilities,
whereas the other deep learning enhancers are less adept at this task for the given input.
When it comes to counteracting the blur by \eg{} restoring the nasal depth structure,
\autoref{fig:ex-synth-samples--2} shows a positive example for most of the deep learning enhancers.
In contrast, \autoref{fig:ex-synth-samples--4} showcases two rather incorrect nose restoration results generated by
DL-Alpha and DL-Gamma,
while the DL-Delta versions and DL-Beta mostly kept the blurred nose instead of overly falsifying the depth data.

The figures show that the
De-M2 enhancer only has
a barely visible impact on the degraded input images,
explaining the
clearly inferior synthetic evaluation results.
De-M8 is more effective, but not enough to fully close all holes, and the depth value of the filled holes is often incorrect, leading to visible artifacts.
Hf-FFA/NFA/L fill most or all of the holes, but due to their design the hole values are mostly incorrect, which results in visible ``smearing'' of the depth values. The same is true for Sp-Hf16/HfU.
Finally, DDRNet-dn's depth range falsification and relative lack of hole-filling can be observed in the figures, which continues for the following real face depth samples.

\subsection{KinectFaceDB Samples}
\label{sec:samples-kinectfacedb}

\begin{figure*}[htb]
	\centering
	\includegraphics[width=\linewidth]{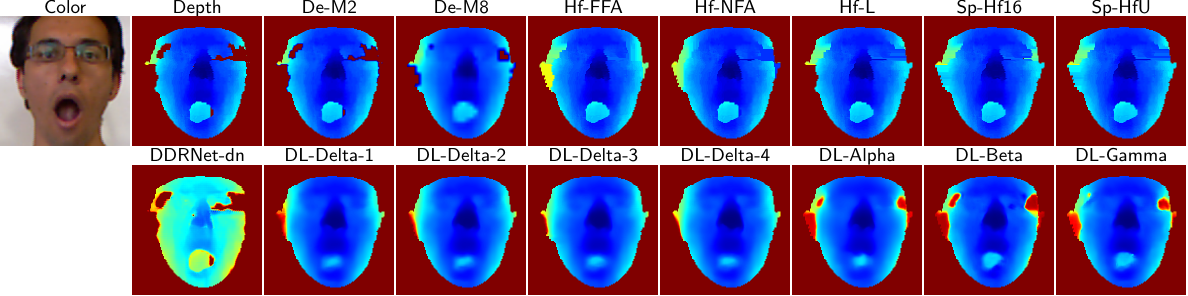}\vspace{-0.2cm}
	\caption{\label{fig:ex-kfdb--0025_s1_OpenMouth} KinectFaceDB sample ``0025\_s1\_OpenMouth''.
	\captionCommonReal{}
	}

\vspace*{\floatsep}

	\centering
	\includegraphics[width=\linewidth]{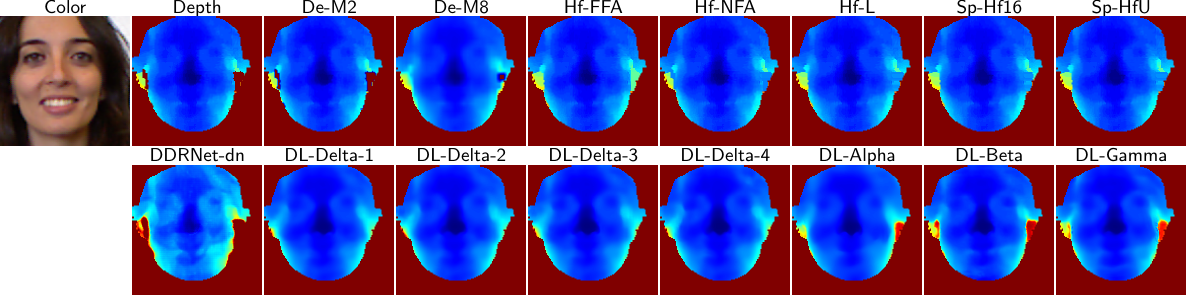}\vspace{-0.2cm}
	\caption{\label{fig:ex-kfdb--0013_s1_Smile} KinectFaceDB sample ``0013\_s1\_Smile''.
	\captionCommonReal{}
	}\vspace{-0.6cm}
\end{figure*}

\autoref{fig:ex-kfdb--0025_s1_OpenMouth} to \autoref{fig:ex-kfdb--0013_s1_Smile} show samples with Kinect v1 input images stemming
from the KinectFaceDB \citep{14-KinectFaceDB}.
The KinectFaceDB images have been preprocessed as described in \autoref{sec:quantitative-real}.

\autoref{fig:ex-kfdb--0025_s1_OpenMouth} shows a subject with open mouth and glasses,
which does not appear to adversely affect the deep learning enhancers,
although the DL-Delta variants and DL-Alpha did visibly shrink the open mouth area.

The other enhancers exhibit the same issues as for the synthetic data:
De-M2 is too ineffective, De-M8 still is insufficient for the larger holes,
Hf-FFA/NFA/L as well as Sp-Hf16/HfU fill the holes but result in visible ``smearing'' artifacts,
and DDRNet-dn falsifies the depth range, with little effect on holes.

\autoref{fig:ex-kfdb--0013_s1_Smile} has a front-facing subject with long hair,
which the DL-Delta variants partially removed as part of the background.
The shown images are however masked to the relevant facial area, so the actual background is not shown.
If the full image area were relevant,
then the degree of falsification for most of the hand-crafted enhancers could often be substantially worse than that of the deep learning enhancers,
since they do not distinguish between the face and the background, meaning that their hole-filling parts can completely alter or ``smear'' pixel values over the entire image.

\subsection{RealSense D435 Samples}
\label{sec:samples-realsense}

\begin{figure*}[htb]
	\centering
	\includegraphics[width=\linewidth]{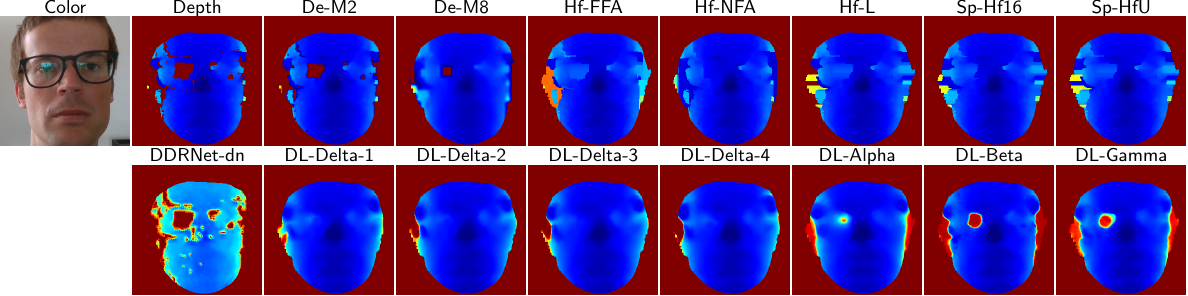}\vspace{-0.2cm}
	\caption{\label{fig:ex-rs--01--2019-09-14T00-31-41} RealSense D435 sample: Frontal with glasses and a large depth hole. Depth: 1280×720, Nearest: 0.34m, ROI: 453².
	\captionCommonReal{}
	}\vspace{-0.0cm}

\vspace*{\floatsep}

	\centering
	\includegraphics[width=\linewidth]{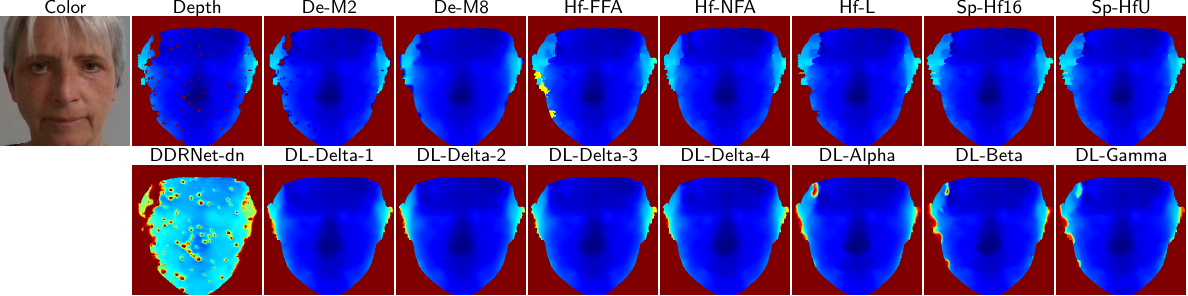}\vspace{-0.2cm}
	\caption{\label{fig:ex-rs--15--2019-09-14T01-40-44} RealSense D435 sample: Frontal with many small depth holes. Depth: 1280×720, Nearest: 0.39m, ROI: 363².
	\captionCommonReal{}
	}\vspace{-0.3cm}
\end{figure*}

\autoref{fig:ex-rs--01--2019-09-14T00-31-41} and \autoref{fig:ex-rs--15--2019-09-14T01-40-44}
show examples for Intel® RealSense™ Depth Camera D435 images recorded as part of this work.
In the upper left of each figure are the two color \& depth source images,
while the other images show the various enhancer outputs.

\begin{itemize}
	\item Color: The visible light RGB color image. All of these shown in this section have been recorded using 1280×720 as the video resolution, whereby the resolution of the depicted square-shaped face ROI can be found in each figure's caption.
	\item Depth:
	The original 16bit output of the RealSense D435 camera is
	aligned with the color video stream,
	and the depth is normalized for a 25cm range starting at the closest detected depth value in the ROI.
    
	These depth images of varying size are scaled to 256×256
	to match the deep learning enhancers' input resolution requirement.
	And while hole-pixels in RealSense depth output is represented with the value 0,
	the deep learning enhancers expect holes to be represented by the maximum value - \ie{} $2^{16}-1=65 535$ for the 16bit depth images. Thus the 0-holes are adjusted by changing their pixel's depth value to the maximum.
	
	The full depth video resolution is denoted in each figure's caption as ``Depth'',
	and the pixel in the ROI with the closest depth value is specified as ``Nearest''.

\end{itemize}

In the frontal \autoref{fig:ex-rs--01--2019-09-14T00-31-41}
a larger depth hole can be seen on the glasses of the subject - here \ia{} the non-Delta deep learning enhancers cannot fully close the hole.
\autoref{fig:ex-rs--15--2019-09-14T01-40-44} showcases a recording containing many small depth holes.
For DDRNet-dn the previously demonstrated problems continue to occur for both of these images.
But while the hand-crafted enhancer results for \autoref{fig:ex-rs--01--2019-09-14T00-31-41} are comparable to the previously shown example figures,
their results in \autoref{fig:ex-rs--15--2019-09-14T01-40-44} illustrate that these enhancers can be suitable for cases with many small depth holes:
Almost all of the hand-crafted enhancers close the holes effectively with mostly correct looking depth data,
although visible artifacts still occur at the face border.
Only De-M2 still appears to be mostly ineffective, despite the small holes.

\section{Conclusion and Future Work}
\label{sec:conclusion-and-future-work}

In this work, we first used PRNet to synthesize suitable ground truth face depth images,
and created synthetically degraded versions thereof to train
seven deep learning enhancer variants with four similar U-Net-like neural network architectures for face-specific depth enhancement.
These were compared against seven hand-crafted general depth enhancers comprising three depth enhancer types from the RealSense SDK,
in addition to a pretrained DDRNet \citep{18-DDRNet} deep learning depth denoiser variant.
The initial quantitative evaluation on the synthetic data indicated that the deep learning approach is effectively enhancing face depth images.
Additional quantitative evaluations on real images from the KinectFaceDB \citep{14-KinectFaceDB}
and from an in-house RealSense D435 dataset
in conjunction with qualitative evaluations support this conclusion.
As part of the qualitative evaluations,
we tested one of the deep learning enhancers in a real-time PAD application using the RealSense D435 \citep{RealSenseD435},
and examined still image enhancer output for the KinectFaceDB, the RealSense D435 camera,
and the synthesized image set.
Overall, the proposed DL-Delta enhancer variants achieved the best results.

Since face depth enhancement for biometric purposes entails a security aspect,
a quantitative and qualitative evaluation was conducted to see whether depth input is falsified to a dangerous degree by the deep learning enhancers,
\eg{} by generating face depth for non-face input.
This does not appear to be the case,
meaning that the enhancers seem sufficiently secure for PAD use.
In this context,
a notable property of the proposed deep learning enhancers is their independence from additional non-depth helper data,
which simplifies the system
and eliminates the potential for depth falsification through such data.

The observed deep learning depth enhancement capabilities lead to the conclusion
that future face depth enhancement research
for facial biometrics
may want to focus on machine learning approaches,
instead of relying on hand-crafted enhancement systems.
Various avenues for future work can be considered, \eg{}:

\begin{itemize}
	\item The network architectures/configurations could be improved.
	AutoML \citep{AML:MSC-2018} variants could be investigated or developed for image-to-image architectures, to create automatically optimized deep learning enhancers.
	For biometric applications such as PAD, it is however important to ensure that the networks do not falsify the output too far - otherwise the enhancement could be more harmful than beneficial.
	Substantial improvements regarding computational performance are presumably achievable.
	\item Different or improved synthesis systems could be utilized.
	Alternatively, a large-scale face depth image dataset with high-quality ground truth images, in addition to recordings made with low-cost depth cameras, could be created.
	Or real data could be combined with synthetic approaches.
	\item While we examined the falsification of non-face depth input, the potential of targeted deep learning adversarial attacks with full image control could still be investigated. Assuming such attacks are effective, they might however not be sensible for real presentation attacks due to the level of noise/defects present with the considered low-cost depth cameras.
\end{itemize}

\section*{Acknowledgements}
This research work has been partially funded by the German Federal Ministry of Education and Research and the Hessian Ministry of Higher Education, Research, Science and the Arts within their joint support of the National Research Center for Applied Cybersecurity ATHENE.

\section*{References}
\AtNextBibliography{\small}
\setcounter{biburllcpenalty}{7000}
\setcounter{biburlucpenalty}{8000}
\printbibliography[heading=none]

\end{document}